\documentclass[10pt,twocolumn,letterpaper]{article}

\usepackage[pagenumbers]{cvpr} % To force page numbers, e.g. for an arXiv version

\definecolor{cvprblue}{rgb}{0.21,0.49,0.74}
\usepackage[pagebackref,breaklinks,colorlinks,allcolors=cvprblue]{hyperref}
\usepackage{graphicx}
\usepackage{amsmath}
\usepackage{amssymb}
\usepackage{booktabs}
\usepackage{multirow}
\usepackage{comment}
\usepackage{xcolor}
\usepackage{pgf-pie}
\usepackage{float}
\usepackage{wrapfig}
\usepackage{etoolbox}
\usepackage{silence}
\makeatletter
\robustify\@latex@warning@no@line
\makeatother
\usepackage{authblk}
\newcommand\philo[1]{{\color{black}#1}}

\def\confName{CVPR}
\def\confYear{2025}

\title{Anticipating Object State Changes in Long Procedural Videos}

\author[1,3,*]{Victoria Manousaki}
\author[1,2,*]{Konstantinos Bacharidis}
\author[1,2,*]{Filippos Gouidis}
\author[3]{Konstantinos Papoutsakis}
\author[1,2]{Dimitris Plexousakis}
\author[1,2]{Antonis Argyros}

\affil[ ]{\{vmanous, gouidis, kpapoutsakis\}@hmu.gr, \{kbach, dp, argyros\}@ics.forth.gr } 
\affil[ ]{} 
\affil[1]{Institute of Computer Science, FORTH }
\affil[2]{Computer Science Department, University of Crete}
\affil[3]{Management Science \& Technology Department, Hellenic Mediterranean University}
\affil[*]{Equal Contribution}

\begin{document}

\twocolumn[{
\maketitle
\begin{center}
    \captionsetup{type=figure}
    \includegraphics[width=\textwidth]{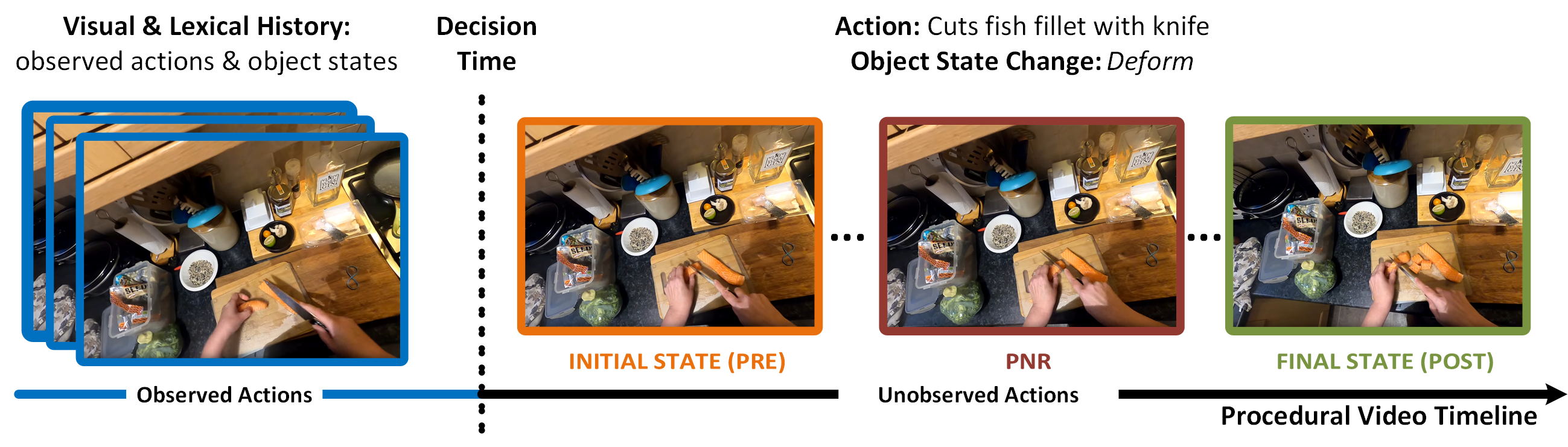}
    \caption{\textcolor{black}{We introduce the new problem of anticipating object state changes in videos of procedural activities, noted OSCA. The decision point in the timeline is placed at the onset of the next anticipated, yet unobserved action. The objective is to predict accurately, at this point, the object state change class that will occur, if any, during the subsequent, yet unobserved action. This involves understanding the dynamics of past and current interactions and how they will affect the object's state.
    An object state change (e.g. deform) refers to a physical and possibly functional change in an object's attributes/properties. It is realized based on the transition from a pre-state (initial) to a post-state (final) occurring at the Point of No Return (PNR) time during an object state-modifying action (e.g. cut fish fillet using a knife).}}
    \label{fig:concept}
\end{center}
}]

\begin{abstract}
In this work, we introduce (a)~the new problem of anticipating object state changes in images and videos during procedural activities, (b)~new curated annotation data for object state change classification based on the Ego4D dataset, and (c)~the first method for addressing this challenging problem. Solutions to this new task have important implications in vision-based scene understanding, automated monitoring systems, and action planning. The proposed novel framework predicts object state changes that will occur in the near future due to yet unseen human actions by integrating learned visual features that represent recent visual information with natural language (NLP) features that represent past object state changes and actions. Leveraging the extensive and challenging Ego4D dataset which provides a large-scale collection of first-person perspective videos across numerous interaction scenarios, we introduce an extension noted Ego4D-OSCA that provides new curated annotation data for the object state change anticipation task (OSCA). An extensive experimental evaluation is presented demonstrating the proposed method's efficacy in predicting object state changes in dynamic scenarios. The performance of the proposed approach also underscores the potential of integrating video and language cues to enhance the predictive performance of video understanding systems and lays the groundwork for future research on the new task of object state change anticipation. The source code and the new annotated data will be made publicly available\footnote{https://projects.ics.forth.gr/cvrl/osca/}.
\end{abstract}    
\section{Introduction}
\label{sec:intro}

When observing human-object interactions, we can effortlessly reason about and anticipate changes in object states
~\cite{bubic2010prediction, pezzulo2008coordinating, bach2014affordance}. 
Imagine, for example, that while preparing the table for a dinner, somebody brings a bottle of wine. Even before opening it, we can infer that in the near future, the bottle will be ``opened'', and glasses will be ``filled''. Recognizing and anticipating object states and their changes is crucial for any entity that interacts with objects because the state of an object significantly affects its physical and functional properties and plays a decisive role in activity understanding, reasoning, and task planning.

\begin{figure}
    \centering
    \includegraphics[width=\columnwidth]{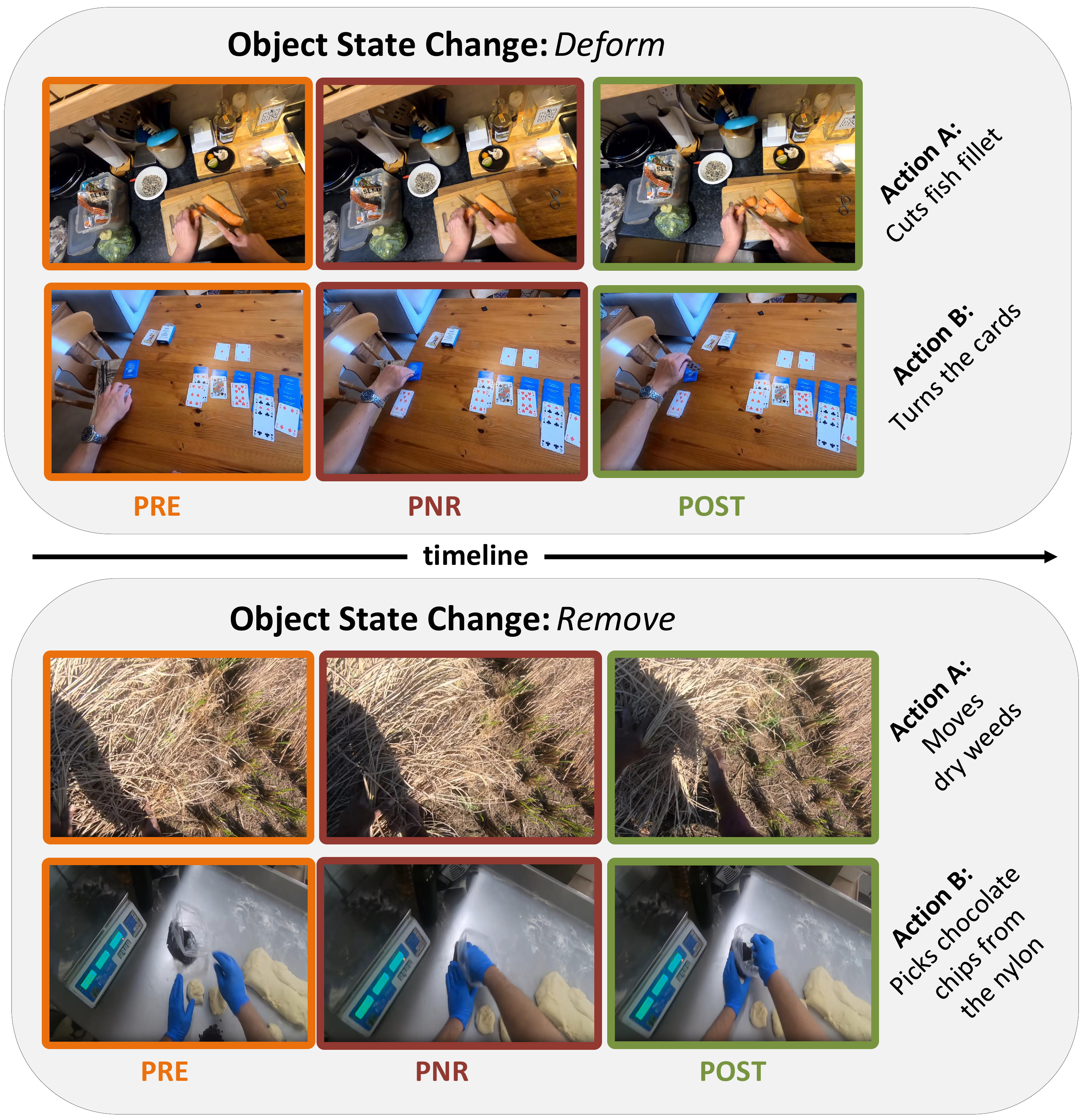}
    \caption{Examples of modifying actions from the ``deform'' and ``remove'' object state change classes represented by a triplet of frames (pre-state, PNR, post-state). Each state change is associated with various actions occurring in diverse environments/scenarios, emphasizing the complexity and challenges introduced in the OSCA problem.} %The PNR time is considered known in our formulation.
    \label{fig:states_inthewild}
    % \vspace{-0.75cm}
\end{figure}

While it is almost effortless for humans, the capability of predicting object state changes still lies beyond the competencies of current AI-powered systems~\cite{zhu2020dark, zellers2019recognition}. 
Understanding object states and their changes in the context of interactions relates to several challenging tasks, such as visual object perception, next-active object prediction, action recognition and anticipation, and object state estimation, that have been well-explored by the research community. Surprisingly, the problem of anticipating object state changes remains undefined and unexplored. However, recognizing and anticipating object state changes would be an important ability of AI-powered agents toward understanding human activities and task planning~\cite{worgotter2009cognitive,10203460}.

The problem of Object State Classification (OSC) is defined as the multi-class recognition of an object's state in a still image~\cite{Isola2015,gouidis2022}, or the initial and the final object states in a video that demonstrates one or more state-modifying actions~\cite{grauman2022ego4d}. The binary object state change classification variant is also related to detecting state change occurrences in a short video clip~\cite{chen2022ego4d,grauman2022ego4d}.
Researchers have only recently started to focus on methods for the representation and understanding of object state changes in videos in the context of state-modifying actions~\cite{saini2023chop,soucek2024genhowto,soucek2022look}, which can also be seen as transformations~\cite{Wang2016b}. Existing benchmarks largely ignore object state changes and focus on traditional types of annotations related to object type, location, or shape, attributes, affordances, and human actions.

In this work, we take one step beyond the Object State Classification and the Action Anticipation tasks by {\sl introducing the new task of Object State Change Anticipation} in videos. OSCA focuses on the multi-class prediction of the state change occurring on an object during the next, yet unseen at inference time, action during a long procedural activity. Specifically, as shown in Fig.~\ref{fig:concept}, at a certain decision point in time that is at the start of the next, yet unobserved action, we aim to predict the object state change class that will occur. The object state change will occur at the ``Point of No Return'' timestamp during the next action~\cite{grauman2022ego4d}. 
% We consider the PNR time known in the proposed formulation.

\begin{figure}
    \centering
    \includegraphics[width=\columnwidth]{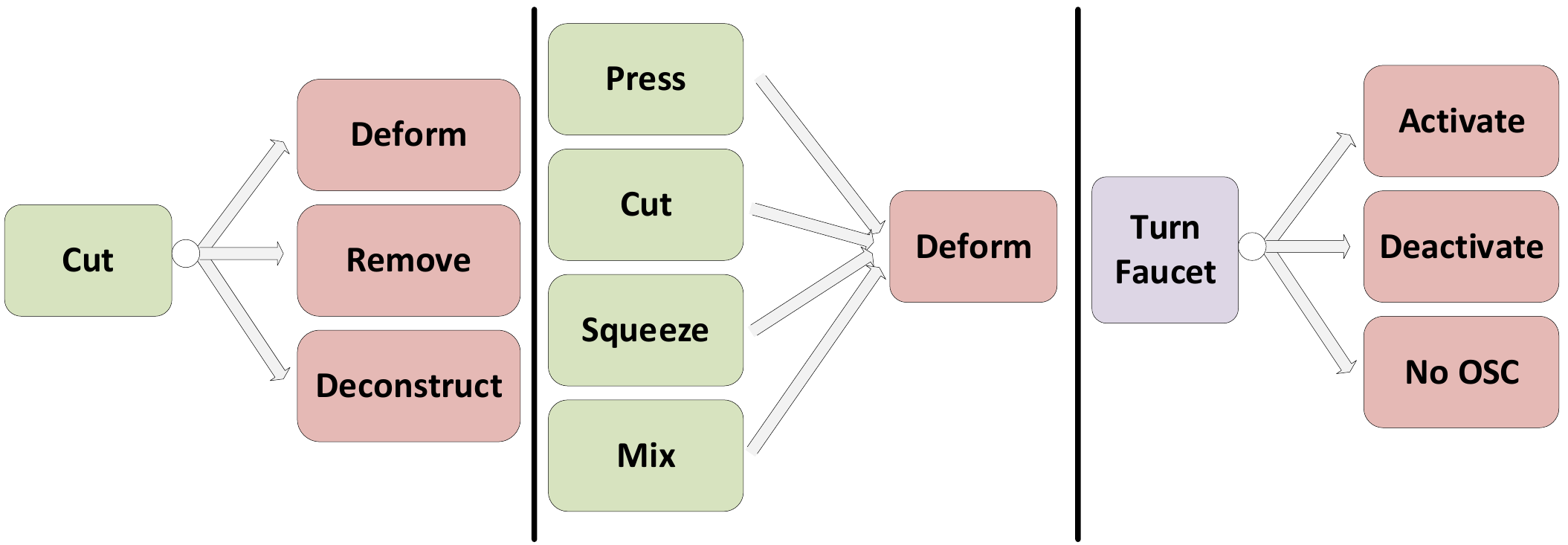}
    \caption{\textcolor{black}{The intricate relation between verb/object/action and object state change. From left to right: one verb may signify different state changes; different verbs might signify the same state change; an action might lead to a variety of object state changes.}}
    \label{fig:diff_states}
\end{figure}

\textcolor{black}{The OSCA task differs significantly from action anticipation, as it focuses on predicting imminent changes in the object's state, if any, without requiring the prediction of the verb and possibly the noun categories associated with the anticipated state-modifying action. Although the OSCA task involves fewer target state change categories than action categories, it remains challenging. A single action may apply to various objects and contexts, resulting in a range of possible state changes, as shown in Fig.~\ref{fig:states_inthewild} and \ref{fig:diff_states}.
In addition, while it might seem that, given a prediction for the next action, determining the next object state change would be straightforward, estimating such predictions in instructional videos is a challenging task, as current benchmarks suggest. Based on the Ego4D leaderboard for short-term action anticipation benchmark}\footnote{\url{https://eval.ai/web/challenges/challenge-page/1623/leaderboard/3910/Overall}} (Top5 mAP: Overall approaches score 7\%, verb and noun combined scores close to 17\%, and noun-only score around 37\% as of 11/2024), it
\textcolor{black}{is clear that a lot of research effort is still needed to devise efficient solutions to action anticipation in procedural activities and consequently their contribution to the estimation of the next state of the action-participating objects.} %In addition, Figure~\ref{fig:diff_states} shows that correctly predicting the anticipated action (verb, noun) does not always ensure accurate determination of the next object state change.}
%We argue that an effective solution to the OSCA task could substantially enhance the performance of action anticipation methods.

To tackle this new task, we introduce a new formulation leveraging on the cumulative history of the textual description of recognized preceding actions and object states, up to the decision point where this prediction is performed. The integration of the former information with visual information concerning the recent past is a key idea to model this historical context effectively. Our approach may also predict that no object state change will occur in the forthcoming action which implies the anticipation of a no-state-modifying action. 
We evaluate the proposed approach toward the newly proposed OSCA problem to establish baseline results for automated change state anticipation in long instructional videos and to also investigate the potential impact and effectiveness of the extracted information in other related vision-based anticipation tasks.
We assess the performance of our proposed method for the OSCA problem and show initial results based on a proposed extension of the popular Ego4D video dataset. We hence build on the large-scale and challenging Ego4D dataset~\cite{grauman2022ego4d} which provides egocentric videos by augmenting the available annotation data with labels for the object state changes based on the initial and the final object states for any state-modifying actions in a subset of videos related to the Hand \& Object Interactions benchmark\footnote{\url{https://ego4d-data.org/docs/benchmarks/hands-and-objects/}}. This results in the Ego4D-OSCA, a variant of the Ego4D dataset that will become available to the community.
Thus, the main contributions are: 
\begin{itemize}
    \item The introduction of the new problem of anticipating an object state change (OSCA) that will occur in the next, yet unseen, action in instructional videos.
    \item The introduction of the Ego4D-OSCA dataset, a new benchmark for evaluating solutions to the OSCA problem based on a subset of the Ego4D video dataset.
    \item The proposal of the first approach to tackle OSCA which integrates visual and language features to model the history of performed actions, object states, and their changes. We also present initial baseline results.
\end{itemize}

\section{Related Work}
\label{sec:related}
Object states capture dynamic aspects of object appearance and/or functionality and are subject to visually perceivable changes, as a result of state-modifying actions. 
They are also known as object fluents related to changeable object attributes~\cite{8237580,Alayrac17,Liu2017}.
%%%AAA: to diatypwsa ligo diforetika parakatw.
Since there is no prior work on the problem of Object State Change Anticipation that we introduce, we 
review the literature on the closely related topics of action and next-active object anticipation, object state classification in images as well as the interplay between object state estimation and action recognition in videos.

\subsection{Object State Classification/Recognition}
\vspace*{0.2cm}\noindent\textbf{Object State Classification in Images}:  
Object states are typically considered as a special subset of ``visual attributes'',  i.e. visual concepts that are related to the physical and functional properties of objects~\cite{Isola2015}. Object states and their changes are perceivable by humans and should be perceivable by AI-enabled agents~\cite{duan2012discovering}. The majority of the attribute classification approaches follow a similar approach to that of object classification by training a convolutional neural network with discriminative classifiers on annotated image datasets~\cite{singh2016end}. 
Few works focus explicitly on state classification~\cite{gouidis2022}, while most of them rely on the same assumptions used for the attribute classification task. 
A prominent direction to tackle this task refers to zero-shot learning. It gained considerable attention in recent years due to its practical significance in real-world applications, mitigating the problem of collecting and learning training data for a very large number of object classes~\cite{xian2018zero}. 
One such prevalent approach involves the utilization of semantic embeddings to represent objects and their attributes in a low-dimensional space~\cite{Wang2018b}. The work in~\cite{gouidis204} leverages Knowledge Graphs (KGs) and semantic knowledge in the context of zero-shot object classification. In a similar vein, the work in~\cite{gouidis2024fusing} combines KGs and Large Language Models (LLMs) to address object-agnostic state classification.
A recent work by~\cite{saini2023chop} focuses on object state recognition based on the compositional generation of novel object-state images, %, also introducing the Chop \& Learn dataset, 
while the method in~\cite{soucek2024genhowto} introduced %the GenHowTo 
a novel conditioned diffusion model that focuses on generating temporally consistent and physically plausible images of actions and object state transformations based on an input image and a text prompt describing the targeted transformation.

\vspace*{0.2cm}\noindent\textbf{Object State Change Estimation \& Action Recognition in Videos}: Object state changes have been considered a meaningful information source in video-based human action understanding and recognition (HAR). 
%%%AAA: Human actions frequently induce modifications in the states of associated objects, thereby influencing the overall dynamics of action sequences. Understanding the causal relationships between actions and object states can enrich our comprehension of action dynamics throughout long complex activities in procedural/instructional videos. Therefore, 
In HAR, object state changes are often considered complementary attributes to the visual representation of actions. These changes are typically derived within the visual domain through the utilization of explicit models for object detection and state estimation~\cite{fathi2013modeling, soucek2022look, liu2017jointly}, or indirect modeling of object states based on general scene changes resulting from action execution~\cite{alayrac2017joint, Bacharidis_2023_ICCV, Bach_Frontiers}.
Several methods exploit object states implicitly to estimate the type of action performed.
The work in~\cite{alayrac2017joint} was among the first to propose a method to automatically discover object states and the associated manipulation actions from videos by leveraging a discriminative clustering framework that jointly models the temporal order of object states and manipulation actions.
The work in~\cite{Liu2017} explored the recognition of object fluents (changeable object attributes) and tasks (goal-oriented human activities) in egocentric videos using a hierarchical model that represents tasks as concurrent and sequential object fluents. 
Moreover, \cite{Ma2018} focuses on understanding human actions within videos by analyzing complex interactions across multiple interrelated objects by recognizing their different state changes. 
In~\cite{soucek2022multi,soucek2022look} a multi-task self-supervised framework is proposed that allows the temporal localization of object state changes and state-modifying actions in uncurated web videos. 

Furthermore,~\cite{xue2024learning} introduced the novel VidOSC approach for understanding object state changes by segmenting object parts related to those changes in videos from an open-world perspective.
A recently proposed framework~\cite{saini2022recognizing} can recognize object-centric actions by relying only on the initial and final object states. The model can also generalize across unseen objects and different video datasets. 
The method proposed in \cite{saini2022disentangling} aims at disentangling visual embeddings that distinctly represent object states alongside identities, enabling effective recognition and generation of novel object-state compositions through a compositional learning framework.
Finally, the InternVideo~\cite{chen2022ego4d} video foundation model was adapted to tackle the tasks of object state change classification and action anticipation in the context of the Ego4D Challenges.

\subsection{Action \& Next-Active Object Anticipation}
%%%AAA: Ayto to section perigrafei poly kathara oles tis ennoies. Wstoso, einai arketa ektetameno kai leptomeres. Epomenws, an xreiastei xwros, ua mporoysame na to meiwsoyme (isws bgazontas ta paradeigmata). 

Action anticipation involves predicting the label of an action that is expected to occur in the future but has not yet been started/observed~\cite{hu2022online,zhong2023survey,wardle2023deep}. This challenge has been studied in both egocentric~\cite{grauman2022ego4d,damen2022rescaling} and exocentric~\cite{sener2022assembly101,huang2024egoexolearn} videos, with the latter becoming increasingly popular in recent years. Short-term action anticipation~\cite{guo2024uncertainty,roy2024interaction,pasca2024summarize} focuses on predicting actions or events in the immediate future, whereas long-term action anticipation~\cite{zhang2024object,mittal2024can} extends to predicting actions or events over a longer period, ranging from several seconds to minutes.

%Uncertainty-aware Action Decoupling Transformer ~\cite{guo2024uncertainty} improves action anticipation by separately predicting verbs and nouns using a two-stream Transformer architecture, where each stream enhances the other, and dynamically combines predictions based on their uncertainties, achieving state-of-the-art results on several benchmarks. Human-object interaction and temporal dynamics are crucial visual cues for egocentric action anticipation, particularly in regions involving objects and human hands. InAViT~\cite{roy2024interaction} model leverages the MotionFormer architecture to incorporate these cues using spatial and trajectory cross-attention.

In the context of human-object interactions in videos, \textit{active objects}~\cite{pirsiavash2012detecting} and \textit{next-active objects}~\cite{FURNARI2017401} refer to specific items that are involved in ongoing or anticipated actions. The active object is the item that a person is currently interacting with within the video. %For example, if someone is shown cutting vegetables, the knife would be the active object as it is directly involved. 
In contrast, the next-active object is the item predicted to be used in the near future, based on the current interaction~\cite{jiang2021predicting, manousaki2022graphing,dessalene2021forecasting}. Although not yet in use, it is likely to become involved in subsequent actions. %For instance, if the person cutting vegetables is expected to reach for a plate to place the cut vegetables on, the plate would be considered the next active object. 
These concepts are crucial in video analysis for understanding and predicting human behavior, as they help anticipate the sequence of actions and interactions within a scene. The concept of next-active object anticipation has also been the subject of the short-term anticipation challenge in~\cite{grauman2022ego4d} and is described as the next object that will be touched by the user (either with their hands or with a tool) to initiate an interaction. Several methods have been proposed in this challenge for the solution of this problem~\cite{thakur2024anticipating,thakur2024leveraging,ragusa2023stillfast,pasca2024summarize}.

\begin{figure}[t]
    \centering
    \includegraphics[width=\columnwidth]{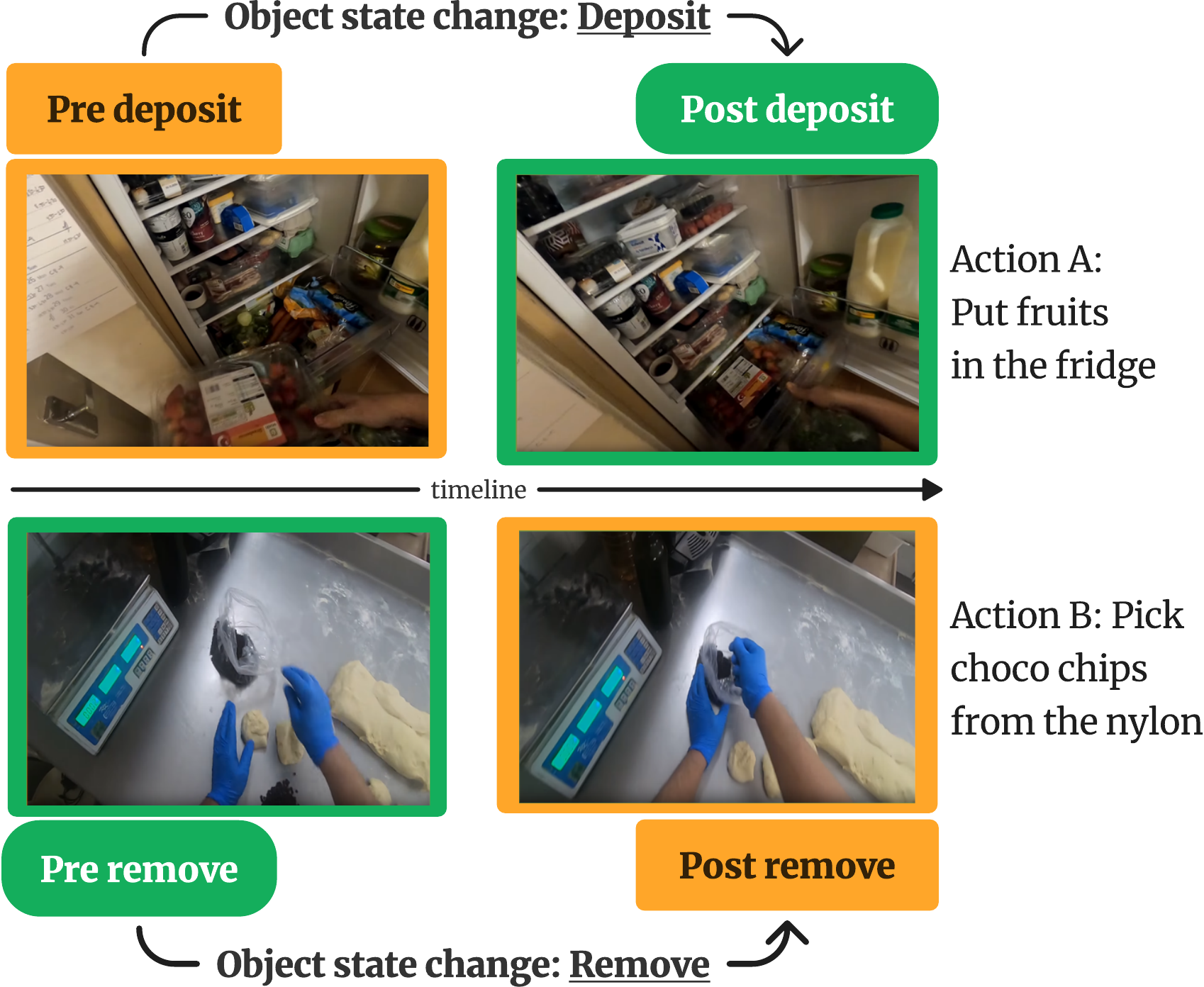}
    \caption{
    Two of the nine object state change super-annotated classes in the Ego4D-OSCA dataset, `deposit' and `remove'. Pre-/post-state labels for these actions are shown as distinct video segments. `Deposit' and `remove' are inverse changes, where pre-deposit matches post-remove, and pre-remove matches post-deposit, indicated by frames and shapes of the same color.}
    %Two of the nine object state change classes that have been super-annotated in the proposed Ego4D-OSCA dataset, ``deposit" and ``remove," are illustrated. Pre-/post-state labels for two state-modifying actions are shown as distinct video segments. ``Deposit" and ``remove'' are inverse state changes, where the pre-deposit matches the post-remove, and the pre-remove matches the post-deposit, as indicated by frames and shapes of the same color.}
    \label{fig:states_annots}
\vspace{-0.2cm}
\end{figure}

Anticipating the state change of an object involves predicting how the object’s condition or form will alter as it becomes involved in an activity, whether it is currently active or will become active in the near future. This process goes beyond merely identifying which object will be used next(\textit{next-active}); it focuses on understanding how the object’s state will evolve during or after its involvement in the interaction. %For example, if someone is going to fry an egg, the egg is initially the next-active object, and the anticipation would involve predicting that its state will change from raw to cooked as it becomes the active object in the frying process. 
This anticipatory process requires analyzing the current interaction and understanding the transformations that occur as objects are used. By predicting these state changes, we gain insights into how objects will behave as they become active or next-active, enhancing the ability to interpret the sequence of events and interactions in a video. 
%Unlike methods that predict the next action or the next-active object, our focus is on predicting the \textbf{state change class} of an object that will be affected in the immediate, yet unobserved, future. For example, consider butter in the oven transitioning from solid to melted. The object in question might be currently involved in the action or may not be used until later. Our approach is unique because it isn’t confined to predicting only active or next-active objects; instead, it anticipates the state change of any relevant object, making it different from standard next-active object prediction methods.}

% Instead, we predict the state change class of the object in the upcoming segment, which might not be the next-active object.

\begin{figure*}[t]
    \centering
    \includegraphics[width=\textwidth]{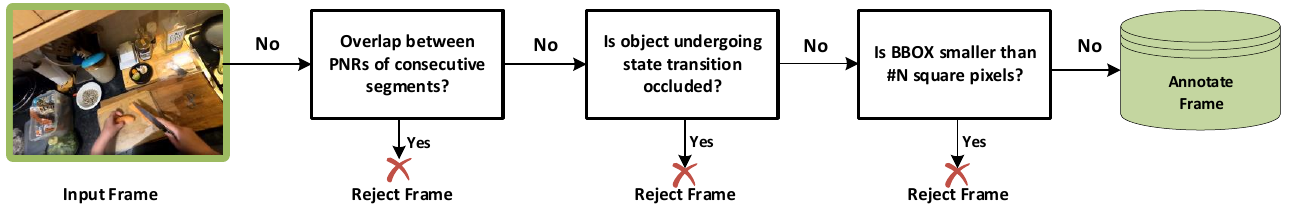}
    \caption{\philo{\textbf{Annotation pipeline:} Occlusions are checked in the pre- \& post-frames. A threshold value for the BBOX area of $N$ square pixels (N=100) for each object annotation. Ego4D-SCOD benchmark data are used to automatically annotate the change states per clip.}}
    %The annotation pipeline. Check occlusions in the pre and post-frames. The default threshold related to bounding box size (Voxel-51~\cite{FiftyOne}) for object classification is employed (100 square pixels). We use the annotation data provided by the SCOD (LTA) benchmark in Ego4D to automatically annotate the change states per video clip.}
    \label{fig:uml}
\vspace*{-.5cm}
\end{figure*}

\section{Ego4D-OSCA Dataset} 

We introduce Ego4D-OSCA as a new partition of the large-scale Ego4D dataset that aims to serve as a benchmark for the assessment of methods for object state change anticipation. The volume and diversity of the Ego4D dataset make Ego4D-OSCA a very challenging dataset for OSCA, as shown in Fig.~\ref{fig:states_inthewild}. Ego4D-OSCA is tailored from the long-term activity (LTA) prediction benchmark, which aims to forecast the sequence of activities that will unfold in future video frames. Due to an ongoing challenge, the official test set for this benchmark has not yet been released, prompting us to re-purpose the validation split as a stand-in test set. 

%TODO:  more details on the cross-check between LTA and SCOD for selecting the videos.

\vspace*{0.0cm}\noindent\textbf{Dataset annotation}: To enrich the LTA benchmark, we integrate object state annotations extracted from the dataset as follows. The original Ego4D dataset does not include annotations for the specific state labels of individual video frames. Instead, annotations about state changes are provided, which relate to entire video segments. Additionally, the dataset includes annotations for bounding boxes and object classes across seven critical frames within each video segment. These frames are temporally centered around the occurrence of the state change that occurs within each video segment. Based on this information, we super-annotate certain critical frames of each video segment with state-related labels as follows. For each video segment, we annotate the initial and final frames as $pre\_X$ and $post\_X$, respectively, where $X$ denotes the label of the state change.
%TODO: add the change state labels + other + justification
Furthermore, in line with the semantic implications of these changes, we establish three pairs of state changes. Each pair is constructed under the premise that the first action is the inverse of the second concerning the resulting state change. For instance, if $X$ and $Y$ represent inverse state changes, then the labels $pre\_X$ and $post\_Y$ are considered samples of identical states. A similar correspondence applies between $pre\_Y$ and $post\_X$. For example, the states $pre\_remove$ and $post\_deposit$ are considered identical, since $remove$ and $deposit$ constitute a pair of inverse state changes. Figure~\ref{fig:states_annots} delineates the specifics for these two super-annotated state change classes. The same condition is true for the state change classes of activate-deactivate and construct-deconstruct. 
The full set of pre- and post- object state pairs that constitute the target set of object state changes appear in the supplementary material.
\philo{It is important to note that there exists a subset of video segments containing actions that do not induce state changes and, therefore, these segments are not considered for annotation. }

%%%TODO more details on the pairs of states --> state changes

By incorporating these detailed annotations, Ego4D-OSCA offers researchers a comprehensive platform to explore and refine methods for anticipating object behavior and activity sequences in egocentric video contexts. 
The Ego4D dataset offers eight distinct state change labels: \textit{activate, deactivate, deposit, remove, construct, deconstruct, deform, and other}. However, we contend that there are actions that do not alter the state of an object. To address this, we propose adding a state change category called ``No Object-State-Change (No OSC)''. This new class will help capture instances where actions occur without affecting the state of an object, thereby providing a more comprehensive framework for understanding and categorizing interactions. 
% \textcolor{red}{Overall, as seen in Fig.~\ref{fig:diff_states} (left, center), a single verb can lead to many different object state changes and vice versa. Also, in Fig.~\ref{fig:diff_states} (right) we observe that actions do not have a one-to-one correspondence with states.}

\begin{table*}[t]
    \centering
    \resizebox{\textwidth}{!}{
    \begin{tabular}{l|c|l|c|rccc} \hline
     \bf Datasets & \bf Modalities\* & \bf \rotatebox{0}{OSC related task}   &\bf \rotatebox{0}{Actions per video} 
& \bf \rotatebox{0}{Samples}& \bf \rotatebox{0}{Obj. State Classes}& \bf \rotatebox{0}{Actions}& \bf \rotatebox{0}{Objects}\\ \hline \hline
     % Isola et al.~\cite{isola2015discovering}& Images &  OS Classification   &2015&N/A & 63.440 & 5 & -&  18   \\ \hline 
     % OSDD~\cite{gouidis2021detecting}   & Images & OS Classification \& Detection   &2021&N/A & 19.000 & 9 & -  & 18    \\ \hline 
     % \hline
     % Alayrac et al.~\cite{alayrac2017joint}& Videos & OS Classification \& Act. Localization  &2017& Single & 630  & 7 & 7 &  5   \\ \hline 
     Fire et al.~\cite{fire2017inferring}& Videos & Detection    & Single & 490 & 17 & - & 13    \\ 
     % Task-Fluent~\cite{liu2017jointly}& Videos & OS Classification  &2017& Single & 809  & 21 & 14 & 25    \\ \hline 
      ChangeIt~\cite{souvcek2022look}   & Videos & Temporal Localization   &Single & 34.428 & - & 44&   -   \\  
      HowToChange~\cite{xue2023learning}   & Video \& Text & Temporal Localization   &Single & 498.475  & 20 &   - &  134   \\ 
      VSCOS~\cite{yu2023video}   & Video & Segmentation   &Single  & 1.905 & 4 & 271 & 124     \\ 
      VOST~\cite{Tokmakov_2023_CVPR}   & Video & Segmentation   &Single & 713 & - &  - & 155    \\ 
    
     Ego4D~\cite{grauman2022ego4d} & Video \& Text & Detection \& Classification & Single & 92.864  & 8 &  - &  478    \\ \hline 
    
     \bf Ego4D-OSCA (Ours) & Video \& Text & \textbf{Anticipation}   &\textbf{Multiple} & 1610  & 9 &  1500 & 477     \\ \hline 
    \end{tabular}
    }    

    \caption{Comparison with other image and video datasets that contain annotations related to object state changes. \*Modalities refer to the available data source for the object state change-related tasks.}
    \label{tab:states_datasets}
    \vspace*{-.5cm}
\end{table*}

\vspace*{0.07cm}\noindent\textbf{Details on the annotation process}: The annotation process for the state transitions is applied to the $pre\_Y$ and $post\_X$ frames in each video segment. Overall, the annotation process consists of the next 4 steps (a schematic representation of the annotation pipeline is shown in Fig.~\ref{fig:uml}). 
First, the PNR moment of the video segment being examined for annotation is compared to the PNR  moment of the segment that has been previously annotated. If the PNR  of the previously annotated segment is located after the PNR of the segment under examination, then the segment under examination is rejected. The reasoning behind this decision has to do with the learning of the segment features related to state transitions.  This alignment of the two PNRs signifies that there is an overlap between the state transition actions of the two segments and therefore the feature learning becomes more challenging.  
Subsequently, it is examined if the object undergoing state change is occluded. If this is the case, the frame is rejected. 
Then, the bounding box area of the object is evaluated, and if it is below 100 square pixels—a threshold empirically chosen and commonly used in annotation tools like Voxel-51—the frame is discarded. Finally, the frame that has passed all the previous checks is annotated with the appropriate state label that pertains to the state transition action.

% \begin{table*}[th]

%     \centering
%     \resizebox{\textwidth}{!}{
%     \begin{tabular}{l|c|c|c|c|c|c|c|c|c} \hline
%      & \bf No OSC & \bf \rotatebox{0}{activate} & \bf \rotatebox{0}{ deactivate} & \bf \rotatebox{0}{construct}& \bf \rotatebox{0}{deconstruct}  & \bf \rotatebox{0}{deposit}& \bf \rotatebox{0}{remove} &   \bf \rotatebox{0}{deform}  & \bf \rotatebox{0}{other}   \\ \hline
%      \bf Train&  2066  & 4017 & 1492 & 4186 & 1773 & 14984 & 15338 & 4400  &   15667    \\ \hline 
%      \bf Test & 1284 & 1888 & 617 & 2289 & 966 & 7613 & 7608 & 2149  &  8715    \\ \hline 
%     \end{tabular}
%     }
%     \caption{Statistics for the Ego4D-OSCA dataset per object state change class. In total, the dataset has 61858 training clips and 31846 testing clips. }
%     \label{tab:states_stats}
 
% \end{table*}

\vspace*{0.0cm}\noindent\textbf{Dataset Statistics}: The proposed \textit{Ego4D-OSCA} dataset is compiled using a subset of the popular, large-scale Ego4D v2 dataset~\cite{grauman2022ego4d} that contains egocentric videos for a large variety of human daily living or work activities.
In Table~\ref{tab:states_datasets} we compare the proposed \textit{Ego4D-OSCA} dataset with existing image and video datasets that also provide annotations related to object states. \textcolor{black}{Ego4D-OSCA contains 61.858 training and 31.846 testing clips.} The target tasks performed using each of the datasets are also noted. More dataset statistics can be found in the supplementary material.

%In the following, we analyze the most similar ones and argue on the necessity to compile a new dataset as a benchmark for the introduced task of Object State Anticipation in videos.

% \begin{figure}
%     \centering
%     \includegraphics[scale=0.27]{images/pie-png.png}
%     \caption{\textcolor{black}{Statistics for the Ego4D-OSCA dataset by object state change class, with 61.858 training and 31.846 testing clips.} }
%     \label{tab:states_stats}
% \end{figure}

%sunolo 61858 clips train
%sunolo 31846 clips test
%sunolo 93704 clips

\section{Object State Change Anticipation - Baseline}
\label{sec:method}

%%%AAA: to parakatw keimeno "myrize" chatgpt, opote aplopoihsa kapoies pio eksezhthmenes lekseis...

\begin{figure*}[t]
    \centering
    \includegraphics[width=\textwidth]{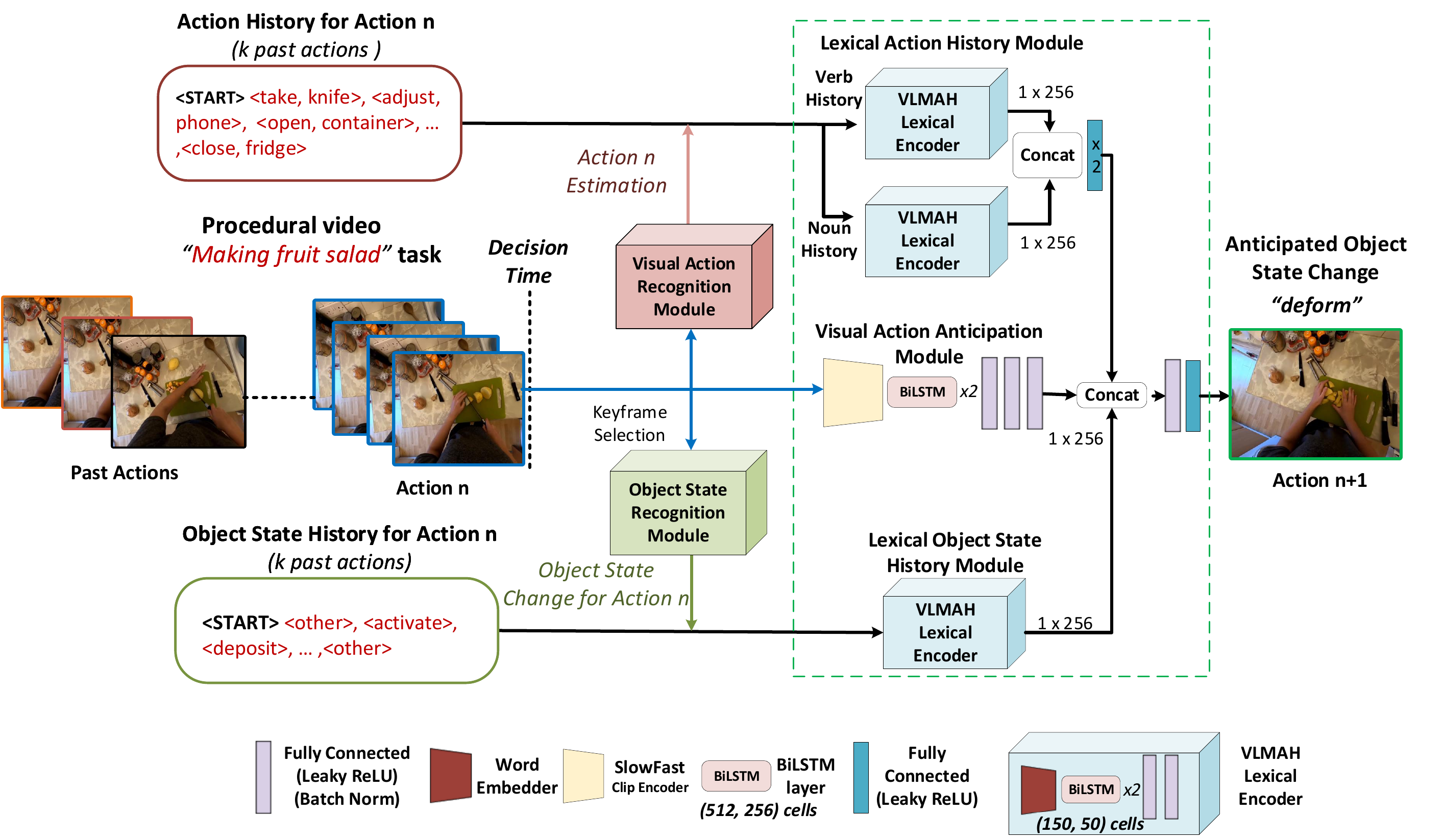}
    \caption{Overview of the proposed baseline framework for the object state change anticipation task. 
The proposed framework anticipates object state changes by integrating real-time visual data and a historical record of past actions and object state changes.}
    \label{fig:Architecture}
    \vspace*{-0.5cm}
\end{figure*}

The proposed framework, depicted in Fig.~\ref{fig:Architecture}, draws inspiration from the efficacy of combining visual and lexical information for semantic action/activity encoding. To achieve this, it adopts a three-stream architecture. Within this design, a visual encoding module is tasked with capturing the visual attributes of ongoing actions, while two lexical-based encoders are employed to extract the semantic nuances from a procedural-oriented representation of past actions and object states. The framework fuses these distinct representations towards the unified objective of anticipating the next object state. This task entails the estimation of the forthcoming state in which the object of interest will reside during the subsequent action. The framework tries to holistically capture the underlying dynamics and contextual intricacies governing object-state transformations across sequential actions by integrating visual and lexical cues.

The design of our framework draws from the recent VLMAH model~\cite{Manousaki_2023_ICCV} that was specifically tailored for the task of {\sl action} anticipation. We augment this architecture by introducing specialized object state history encoding modules. Additionally, we redesign the action history module to facilitate disjoint encoding, capturing both the motion motifs in actions (verbs) and the transitions of objects-in-use (nouns) between actions. This refined architecture enables a more nuanced representation of the sequential dynamics between actions and object states, empowering the framework to achieve enhanced performance in the task of next-state anticipation within dynamic environments.

As illustrated in Fig.~\ref{fig:Architecture}, the proposed framework consists of two primary components: (a)~the current action and object state estimation module, and (b)~the object state anticipation module, depicted within the thin-dotted rectangle. Our contribution resides in (a)~the conceptualization of this framework and (b)~the development of the object state anticipation module. Concerning the latter, it encompasses some constituent components.

\vspace*{0.cm}\noindent \textbf{Visual Encoder}: For this module, we employ a lightweight visual encoder consisting of a single-branch bidirectional long short-term memory (BiLSTM) component followed by a multi-layer perceptron (MLP). We selected this simplified design for the visual encoder based on the objective of temporally encoding the enduring relationships among encoded short-term segments extracted from the input video. Our model relies on an external pretrained human action recognition model, such as SlowFast~\cite{feichtenhofer2019slowfast} or TSN~\cite{TSN} to provide encodings of short-term spatio-temporal dependencies between the frames inside a single segment.

\vspace*{0.cm}\noindent \textbf{Action \& State History Encoders}: As illustrated in Fig.~\ref{fig:Architecture}, both encoders exploit the model design of the lexical encoder of the VLMAH model~\cite{Manousaki_2023_ICCV}, which follows a NLP neural network design consisting of BiLSTM and MLP components. The decision to employ a simple NN for encoding the history, instead of utilizing LLMs was motivated by several factors. Firstly, the computational efficiency of LLMs such as GPT- or LLaMA, often entails significant resource requirements for training and inference~\cite{wan2023efficient}, whereas a simpler neural network architecture mitigates computational overhead. Secondly, LLMs are pre-trained on general text corpora and may not capture the domain-specific nuances inherent in the textual data related to action histories and object states. Additionally, the simplicity of the chosen architecture facilitates interpretability, data efficiency, and customization, affording greater control over the model's behavior and adaptation to the task's requirements.

\vspace*{0.cm}\noindent \textbf{Learning Objective}: The objective for training the model was exclusively focused on evaluating the anticipated state estimate. This deliberate choice stemmed from the aim to prioritize the accurate prediction of object states, which was the study's primary objective. This objective was formulated using the cross-categorical entropy loss, which is well-suited for multi-class classification tasks, such as predicting object states across different categories:
\vspace{-1.7mm}
\begin{equation}
\mathcal{L} = -\frac{1}{N} \sum_{i=1}^{N} \sum_{c=1}^{C} y_{i,c} \log(\hat{y}_{i,c}),
\end{equation}
where $N$ is the number of samples in the dataset, $C$ is number of object state categories, $y_{i,c}$ is the ground truth next state label for the object-in-use in the current action sample $i$ and $\hat{y}_{i,c}$ is the predicted next state probability.

During training, the proposed framework leverages oracle action and state detectors to provide the action and state history, respectively, for each clip in the dataset (see Fig.~\ref{fig:Architecture}). These detectors estimate the current action and object state observed in the clip, serving as ground truth annotations for training purposes. However, it is important to note that for inference toward real-world applications, there is a requirement for current action and object state recognition models to provide input to the framework. 
Consequently, our model is solely tasked with the learning objective of next-object anticipation, focusing exclusively on predicting the future state of the object. By decoupling the training and inference phases in this manner, the model can effectively learn the dynamics of object-state transitions without the added complexity of simultaneously predicting the current action.

\section{Implementation, Experiments and Results}
\label{sec:exp}

\vspace*{0.cm}\noindent{\bf Implementation Details:} The proposed state anticipation model (dotted rectangle in Fig.~\ref{fig:Architecture}) is trained on a single NVIDIA TITAN GPU using the Adam optimizer, a batch size of 32,  a learning rate of $1e-4$,  without any temporal augmentations (clip or frame cropping). Short-term associations between neighboring segments of an input video are represented using the pre-extracted SlowFast frame level features from Ego4D. Regarding the selection of the pre- and the post-state keyframes that are introduced in the object recognition model, we exploit the PNR annotations of Ego4D, which correspond to the first frame in each clip when the state change/transition is visible. We should note that in real-world inference, the action and state lexical histories in the proposed anticipation model will be populated by existing action recognition and object state estimation models trained on the respective data of the task.
%%%AAA: Kapou edw (mallon) as mpei mia mikri estw anafora sta evaluation metrics. 
%%%KKBach: addressed

\vspace*{0.cm}\noindent{\bf Evaluation metrics:} The evaluation of all examined models was conducted using top-1/5 mean accuracy, and F-score, following standard practices in the relevant literature.

\subsection{OSCA Results}

In Table~\ref{tbl::state_ant}, we compare variants of an object state anticipation model to highlight the impact of incorporating lexical histories of past actions and object states on the anticipation performance. The vision-only model (VID-A) only relies on the visual representation of the current action. We observe modest performance levels. 

\noindent\textbf{OSCA under ideal action and state recognition}:
When ground-truth lexical histories of past actions are introduced through an oracle recognition model (VNLP (O-Action)), a slight performance improvement indicates the potential benefit of contextual action information. Notably, incorporating lexical histories of past object states from an oracle recognition model (VNLP (O-State)) leads to significant performance gains, that highlight the importance of considering object state dynamics in anticipation tasks. 
Further improvements are observed when both lexical histories are integrated into the model (VNLP (O-Action, O-State)), demonstrating the synergistic effect of leveraging contextual information from actions and object states. Overall, the low anticipation scores highlight the inherently challenging nature of the task and the intricacy of the dataset scenarios that pose significant challenges for anticipation models.

\noindent\textbf{OSCA under actual action and state recognition}:
We conducted experiments where the oracles (action recognizer, object's current state estimator), were substituted with existing baseline models. In this setting, the output of these models is utilized to populate the action and state histories that OSCA utilizes. We employed the following models: (a) the well-established SlowFast~\cite{feichtenhofer2019slowfast} model for action recognition; (b) the object-agnostic state classification method proposed by \cite{gouidis2022} as the current state classifier, with minor modifications, as described below. 

\noindent{\sl Action recognizer}: Regarding the action classification module, we fine-tuned the SlowFast model~\cite{feichtenhofer2019slowfast} on a subset of the original Ego4D dataset. Specifically, we obtained the train/validation/test splits using the training set provided for the (LTA) long-term anticipation task of Ego4D based on the $60/20/20$ split scheme. The adaptation of the LTA data to the action recognition task resulted in $5754$ action classes and a total of $\approx65K$ video clips (with a mean of $\approx10.7$ sample clips per action class). SlowFast achieved $12.86\%$ Top-1 and $33.69\%$ Top-5 accuracy for the task of current action recognition. This low performance can be attributed to the extensive number of action classes and limited samples per class, as well as in Ego4D's inherent motion and appearance similarities across different actions, e.g. \textit{take cup - take bottle, tie string - tie rope}.
\begin{table}[t]
\centering
\small
\begin{tabular}{p{4.0cm}|c|c}
\hline
\bf Model & \bf Top@1/5 mAcc & \bf F1-score  \\
\hline
VID-A & 23.93 / 89.10\% & 11.74\% \\
\hline
VNLP (O-Action) & 25.59 / 83.06\% & 24.62\% \\
VNLP (O-State) & \textbf{40.07} / 90.83\% & 33.57\%\\
VNLP (O-Action, O-State) & 39.20 / 89.76\% & \textbf{37.12}\% \\
\hline
VNLP (Action~\cite{feichtenhofer2019slowfast}) & 23.04 / 81.31\% & 22.09\% \\
VNLP (State~\cite{gouidis2022}) & \textbf{32.72} / 92.16\% & 21.78\% \\
VNLP (Action \cite{feichtenhofer2019slowfast}, State \cite{gouidis2022}) & 29.42 / 94.65\% & \bf 26.29\% \\
\hline
\end{tabular}

\caption{OSCA performance for various model configurations (O-: Oracle recognizer, VID-A: vision-only state anticipation model).}
\label{tbl::state_ant}
% \vspace{-0.3 cm}
\end{table}

\noindent{\sl Object state recognizer:} Additionally, for the object-agnostic state history, we adapt the model of \cite{gouidis2022}. This model relies on the outputs of two distinct state classifiers. Each classifier receives the first (pre) or the last (post) frame of each video segment as input to predict the object state label for the respective frame. The prediction of the state-change label for the video segment considers both outputs and is derived based on the following rules.
If the object state predictions are $pre\_X$ and $post\_X$, respectively, the inferred state change for the video segment is denoted as $X$. Conversely, if the classifiers predict $pre\_X$ and $post\_Y$, where $X$ and $Y$ are distinct and represent inverse state changes, it is concluded that no state change has occurred. Finally, if neither of the above conditions is met, the prediction of the state change defaults to the output of the second classifier; that is, if the prediction is $post\_Y$, the state change for the video segment is identified as Y. For example, if the predictions of the two classifiers are $pre\_activate$ and $post\_activate$ the prediction of state change would be $activate$. Likewise, if the predictions are $pre\_activate$ and $post\_deactivate$ the prediction of the state change would be that of no change. This object-agnostic state recognizer showcased $25.4\%$ mean state recognition accuracy.

In our experiments, replacing oracles with realistic recognizers to populate the action and state history buffers that are considered by the baseline OSCA model, we observed a significant accuracy drop (last block of lines in Table~\ref{tbl::state_ant}). This accuracy difference underscores the critical role of precise recognition of the current action and object state for effective anticipation of near-future object states within dynamic environments.
\textcolor{black}{Given the class imbalance in the proposed Ego4D-OSCA data set, the F1 score is a more appropriate performance measure. This metric considers both precision and recall while remaining insensitive to the true negatives of majority classes, unlike accuracy, which can be biased toward the majority class. Based on this rationale, the reported F1-scores in Table~\ref{tbl::state_ant} indicate that a model combining both action and state history (past context) may be more effective for the object state anticipation task.}

% \textcolor{red}{Additionally, the F1-scores across the different model versions indicate that in realistic conditions a model that combines both action and state history (past context) is potentially more suitable for the object state anticipation task, since it is able to more robustly handle the imbalanced representation of objects states that are present in real-world activities.}  

\begin{table}[h]
\centering
\begin{tabular}{ll}
 \hline

\bf Noise (Action, State)  & \bf  Top@1/5 mAcc \\
\hline
(0\%, 0\%) (Oracle)   & \bf  35.60 /  \bf 88.14\% \\
(25\%, 25\%)   &   30.46 / 84.42\% \\
(50\%, 50\%)   &   26.00 / 81.75\%  \\
(75\%, 75\%) &   22.48 / 78.09\% \\
\hline

\end{tabular}

\caption{The robustness of the object state change anticipation model is tested to the recognizer performance variability. }
\label{tbl::noise}
% \vspace{-.5cm}
\end{table}

% \begin{table*}
% \parbox{.5\linewidth}{
% \small
% \centering
% \begin{tabular}{ll}

% \hline
% \bf Model  &  \bf Top@1/5 mAcc \\
% \hline
% VID-A  & 23.93 / 89.10\%\\
% % \hline
% VNLP (O-Action) & 25.59 / 83.06\% \\
% VNLP (Action~\cite{feichtenhofer2019slowfast}) & \% \\
% \hline
% VNLP (O-State) &  34.35 / 85.92\% \\
% VNLP (State~\cite{gouidis2022}) & 30.23 / 91.80\% \\
% \hline
% VNLP (O-Action, O-State) & \textbf{35.60} / 88.14\%   \\
% VNLP (O-Action, State~\cite{gouidis2022}) & 28.56 / \bf 92.24\%   \\
% VNLP (Action~\cite{feichtenhofer2019slowfast}, State~\cite{gouidis2022}) &   \%  \\
% \hline
% \end{tabular}
% \caption{Object state change anticipation performance for various model configurations (O-: Oracle recognizer, VID-A: vision-only state anticipation model).}
% \label{tbl::state_ant}
% }

% \hfill
% \parbox{.45\linewidth}{
% \small
% \centering
% \begin{tabular}{ll}
%  \hline

% \bf Noise (Action, State)  & \bf  Top@1/5 mAcc \\
% \hline
% (0\%, 0\%) (Oracle)   & \bf  35.60 /  \bf 88.14\% \\
% (25\%, 25\%)   &   30.46 / 84.42\% \\
% (50\%, 50\%)   &   26.00 / 81.75\%  \\
% (75\%, 75\%) &   22.48 / 78.09\% \\
% \hline

% \end{tabular}
% \caption{Object state change anticipation model robustness to recognizer performance variability. }
% \label{tbl::noise}
% }

% \end{table*}

\vspace{-0.1cm}\subsection{Object State \& Action Recognition Impact}

To further demonstrate the impact of the current action and object state recognizer accuracies on the object state change anticipation task, we conducted experiments that hypothesized recognizers of different accuracy. In this experimental setup, we uniformly introduce noise, representing erroneous estimations, to both the action and state histories, since in the inference stage of the proposed framework, these histories would need to be populated by the outputs of the respective recognizers.

Table~\ref{tbl::noise} presents the results obtained under three varying levels of label noise (rows 2-4), contrasted to the outcomes achieved when employing ground truth labels (where the noise level is 0\%). The noise levels correspond to the rate of erroneous estimates generated by the recognizers, i.e., $25\%$ corresponds to a recognizer with $75\%$ mAcc. As it can be verified based on the obtained results, the performance of the state anticipation task is influenced by the recognizer's accuracy, demonstrating an approximate $4-5\%$ reduction in OSCA accuracy for every $25\%$ decrease in object state and action recognition accuracy. Notably, despite substantial declines in state and action recognition performance, the anticipation model exhibits only a marginal decrease in performance. This finding can be attributed to the compensatory capability of the visual component of the model, which effectively accommodates dynamic and previously unseen sequences of action and state histories.

\section{Conclusions}

This paper introduced the new problem of object state change anticipation during procedural activities. We proposed a novel framework that integrates lexical histories of past actions and object states with recent visual information to enhance anticipation accuracy in vision-based models. By fusing long-term semantic and recent visual information, our framework demonstrates notable improvements in anticipation accuracy, underscoring the importance of contextual understanding in dynamic environments. To validate our approach, we augmented the Ego4D dataset forming a specialized subset noted Ego4D-OSCA. 
% that could benefit from using LLMs for better semantic understanding, albeit with increased computational costs. 
Future work will explore the applicability of LLMs as a replacement for the NLP processing component of the proposed framework, for leveraging their enhanced semantic understanding and in-context learning abilities. We also plan to explore zero-shot settings to enable anticipation of state changes involving novel, previously unseen, objects or actions.

% \section{Final copy}

% You must include your signed IEEE copyright release form when you submit your finished paper.
% We MUST have this form before your paper can be published in the proceedings.

% Please direct any questions to the production editor in charge of these proceedings at the IEEE Computer Society Press:
% \url{https://www.computer.org/about/contact}.
% 
{
    \small
    \bibliographystyle{ieeenat_fullname}
    \bibliography{main}
}

\clearpage
\setcounter{page}{1}
\maketitlesupplementary

This supplementary material aims to provide a detailed analysis of the proposed \textit{Ego4D-OSCA} dataset. 
In Section~\ref{sec:compare}, we compare the proposed video dataset with existing image and video datasets proposed to support vision-based tasks related to object state change understanding. Section~\ref{sec:analysis} provides an analysis of the complexity of the proposed dataset in the context of the newly introduced object state anticipation task. 
In that direction, we highlight its challenging nature that arises from the characteristics and underlying associations among the actions, the activities, the objects, and the annotated object states.
Finally, in Section~\ref{Sec: ObjectState_general} additional sample images retrieved from the proposed dataset are shown to emphasize the complexity and context variability of object state change and action classes. 

%-------------------------------------------------------------------------
\section{Comparison of datasets related to object state changes}
\label{sec:compare}

%In this section, a detailed description of the proposed \textit{Ego4D-OSCA} dataset is reported. 
The proposed  
\textit{Ego4D-OSCA} dataset is compiled using a subset of the popular, large-scale Ego4D v2 dataset~\cite{grauman2022ego4d} comprising egocentric videos for a large variety of human daily living or work activities.

In Table~\ref{tab:states_stats} we compare the proposed \textit{Ego4D-OSCA} dataset with existing image and video datasets that also provide annotations related to object states. The target tasks performed using each of the datasets are also noted.
In the following, we analyze the most similar ones and argue on the necessity to compile a new dataset as a benchmark for the introduced task of Object State Anticipation in videos.

% Here we will add also the histogram of the number of states per video, and explain the complexity (i.e. 880 states in video 14 for example (video name is 1e5bd816-e1dd-43d3-8709-42c83114dc7c).

\begin{table*}[th]
    \centering
    \resizebox{\textwidth}{!}{
    \begin{tabular}{|l|c|l|l|c|r|c|c|c|} \hline
     Datasets & \bf Modalities & \bf \rotatebox{0}{Task}   &\bf \rotatebox{0}{Year}&\bf \rotatebox{0}{Actions per video} 
& \bf \rotatebox{0}{Samples}& \bf \rotatebox{0}{Obj. State Classes}& \bf \rotatebox{0}{Actions}& \bf \rotatebox{0}{Objects}\\ \hline \hline
     Isola et al.~\cite{isola2015discovering}& Images &  OS Classification   &2015&N/A & 63.440 & 9 & -&  18   \\ \hline 
     OSDD~\cite{gouidis2021detecting}   & Images & OS Classification \& Detection   &2021&N/A & 19.000 & 9 & -  & 18    \\ \hline 
     \hline
     Alayrac et al.~\cite{alayrac2017joint}& Videos & OS Classification \& Act. Localization  &2017& Single & 630  & 7 & 7 &  5   \\ \hline 
     Fire et al.~\cite{fire2017inferring}& Videos & SC Object Detection    &2017& Single & 490 & 17 & - & 13    \\ \hline 
     Task-Fluent~\cite{liu2017jointly}& Videos & OS Classification  &2017& Single & 809  & 21 & 14 & 25    \\ \hline 
      ChangeIt~\cite{souvcek2022look}   & Videos & OSC Temporal Localization   &2022&Single & 34.428 & - & 44&   -   \\ \hline 
      HowToChange~\cite{xue2023learning}   & Video \& Text & OSC Temporal Localization   &2023&Single & 498.475  & 20 &   - &  134   \\ \hline 
      VSCOS~\cite{yu2023video}   & Video & SC Object Segmentation   &2023&Single  & 1.905 & 4 & 271 & 124     \\ \hline 
      VOST~\cite{Tokmakov_2023_CVPR}   & Video & SC Object Segmentation   &2023&Single & 713 & - &  - & 155    \\ \hline 
      MOST~\cite{tateno2024learning}   & Video & OS Classification   &2024 &Multiple & 61 & 60 &  - & 6    \\ \hline 
     \hline
     Ego4D~\cite{grauman2022ego4d} & Video & SC Object Detection \& Classification &2022& Single & 92.864  & 8 &  - &  478    \\ \hline 
     \hline
     \bf Ego4D-OSCA & Video & \textbf{OSC Anticipation}   & 2024 &\textbf{Multiple} & 1498  & 9 &  5754 & 475     \\ \hline 
    \end{tabular}
    }    
\vspace*{0.3cm}
    \caption{Comparison with other image and video datasets that contain annotations related to object state changes. Note that in \textit{Ego4D-OSCA}, a sample refers to a video of an entire activity, which might consist of multiple actions.}
    \label{tab:states_stats}
 
\end{table*}

\noindent \textbf{Ego4D~\cite{grauman2022ego4d}:} The Ego4D dataset is among the largest to date, encompassing an extensive collection of videos captured in a wide variety of environments. A subset of this dataset has been utilized for object state detection and classification, as shown in the second-to-last row of Table~\ref{tab:states_stats}. This subset consists of 92,864 short videos, each featuring a single state-modifying action.

The Ego4D dataset is related to three key challenges concerning visual object state change understanding. Firstly, the Ego4D SCOD (State Change Object Detection) challenge\footnote[1]{\href{https://eval.ai/web/challenges/challenge-page/1632/overview}{Ego4D State Change Object Detection Challenge}} focus on the bounding box-based detection of the object that undergoes a state change in an action segment (short video clip).
The State Change Classification task is defined as the multi-class classification of object state changes in a video clip where a state-modifying action occurs, for example, identifying that the state of a cup has changed from filled to empty. Second, the binary state change classification variant is realized as an Ego4D challenge\footnote[2]{\href{https://eval.ai/web/challenges/challenge-page/1627/overview}{Ego4D Object State Change Classification Challenge}}, with the aim to detect whether a state change was performed or not in an action segment (video clip). 
Moreover, the Ego4D State Change Localization  challenge\footnote[3]{\href{https://eval.ai/web/challenges/challenge-page/1622/overview}{Ego4D Object State Change Temporal Localization Challenge}} involves pinpointing the exact frames in the video where the state change occurs. Accurate localization is crucial for understanding the precise timing and context of the state transitions within the egocentric video perspective.
These challenges are designed to advance the understanding and development of AI models in recognizing and interpreting state changes in dynamic and realistic scenarios captured from a first-person viewpoint.
Finally, a series of workshops in major conferences have been organized based on the Ego4D dataset and related tasks/benchmarks, such as 
\href{https://ego4d-data.org/workshops/eccv22/}{2nd International Ego4D Workshop @ ECCV 2022
},
\href{https://ego4d-data.org/workshops/cvpr22/}{1st Ego4D Workshop @ CVPR 2022
} and the 
\href{https://cvpr.thecvf.com/virtual/2023/workshop/18537}{Joint 3rd Ego4D and 11th EPIC Workshop on Egocentric Vision @ CVPR2023}.
%------------TABLE------------------------
\begin{table*}[th]

    \centering
    \resizebox{\textwidth}{!}{
    \begin{tabular}{l|c|c|c|c|c|c|c|c|c} \hline
     & \bf No OSC & \bf \rotatebox{0}{activate} & \bf \rotatebox{0}{ deactivate} & \bf \rotatebox{0}{construct}& \bf \rotatebox{0}{deconstruct}  & \bf \rotatebox{0}{deposit}& \bf \rotatebox{0}{remove} &   \bf \rotatebox{0}{deform}  & \bf \rotatebox{0}{other}   \\ \hline
     \bf Train&  2066  & 4017 & 1492 & 4186 & 1773 & 14984 & 15338 & 4400  &   15667    \\ \hline 
     \bf Test & 1284 & 1888 & 617 & 2289 & 966 & 7613 & 7608 & 2149  &  8715    \\ \hline 
    \end{tabular}
    }
    \caption{Statistics for the Ego4D-OSCA dataset per object state change class. In total, the dataset has 61858 train and 31846 test clips. }
    \label{tab:states_number_stats}
 
\end{table*}
%------------END OF TABLE------------------------

\noindent\textbf{Comparison:} The Ego4D-OSCA dataset comprises long videos of sequential state-modifying actions that correspond to any of the nine classes of object state changes: \textit{deposit, remove, construct, deconstruct, activate, deactivate, deform, other, and no-state-change}. A distribution of the samples across the 9 object state labels is presented in Table~\ref{tab:states_number_stats} and Fig.~\ref{tab:states_stats}. In contrast, the current subsets of Ego4D used for detecting and classifying object states (Ego4D SCOD \& OSCC benchmark) comprise short videos, each depicting a single action. 
This renders the subsets unsuitable for addressing the problem of anticipating object state changes \textcolor{black}{in procedural videos that comprise consecutive actions under the same scenario (activity).} 
\vspace*{0.2cm}
%Following, we will outline the most relevant and up-to-date datasets that are similar to ours.

\begin{figure}
    \centering
    \includegraphics[scale=0.27]{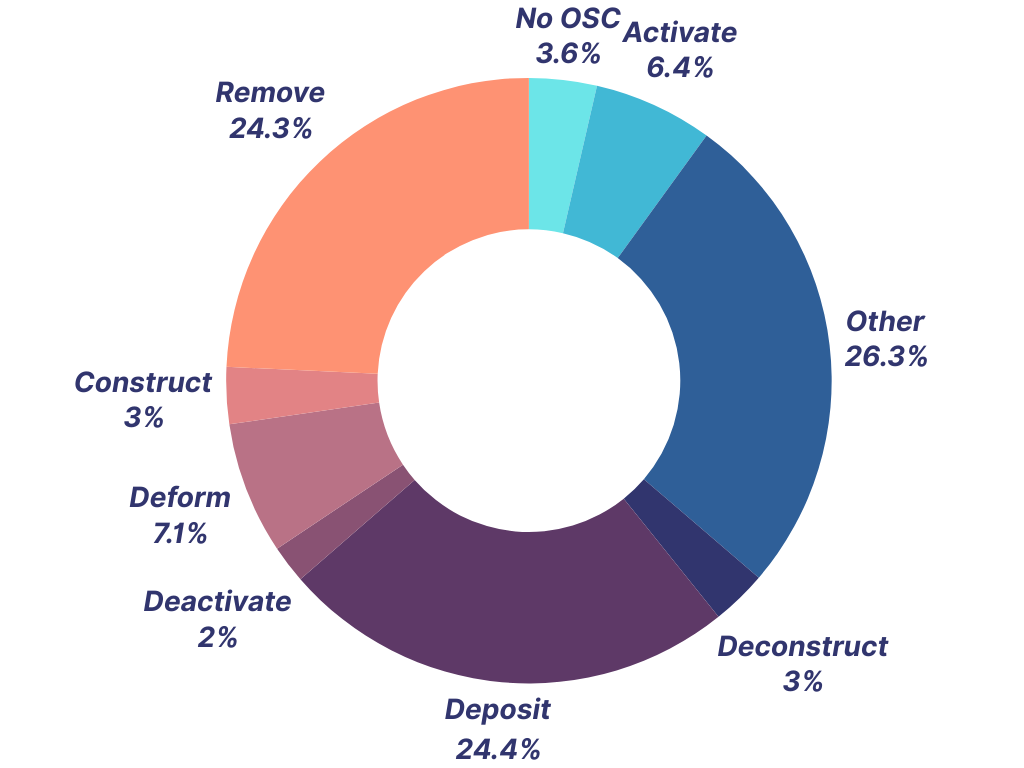}
    \caption{Statistics for the Ego4D-OSCA dataset per object state change class.}%, with 61.858 training and 31.846 testing clips.} }
    \label{tab:states_stats}
\end{figure}

%%%% This dataset contains initial state, action, end state labels. are thesee considered Obj state classes?
\noindent\textbf{ChangeIt~\cite{souvcek2022look}:} The ChangeIt dataset comprises unedited videos sourced from YouTube and automatically generated labelling of actions. The designated tasks for analysis on this dataset involve the identification and temporal localization of the initial state, end state, and state-modifying action in a video. A set of 44 state-changing actions is provided, each demonstrated in approximately $15$ videos on average. In total, there are $34,428$ videos with an average duration of 4.6 minutes.
% The designated tasks for analysis involve the temporal localization of states and action frames.

\noindent\textbf{Comparison:} The newly introduced Ego4D-OSCA dataset comprises sequential videos featuring actions, some of which may involve state changes, while others may not. In contrast, the ChangeIt dataset consists of single-action clips, therefore one object state change is performed per clip. Ego4D-OSCA offers a wide range of scenarios without imposing any limitations and encompasses video durations spanning from minutes to hours. Conversely, the ChangeIt dataset confines scenarios to irreversible actions, aiming to eliminate instances where two actions return an object from an initial state to the same initial state via an intermediate state. Additionally, it excludes videos exceeding 15 minutes in length. 
\vspace*{0.2cm}

\textcolor{black}{
\noindent\textbf{MOST~\cite{tateno2024learning}:} The newest dataset related to object states in videos addresses the problem of temporal segmentation of multi-label object states. It includes manually collected instructional videos from YouTube, covering six object categories: apple, egg, flour, shirt, tire, and wire, each with around 10 annotated object states. These states represent common appearances or conditions an object may take. Annotators marked the time intervals when specific object states were visible, resulting in a dataset of 61 fully annotated videos with a total duration of 159.6 minutes. Unlike other datasets, such as ChangeIt, which focuses only on state transitions, MOST captures diverse object states, even if those are not tied to specific actions, offering a comprehensive benchmark for object state recognition.}

\textcolor{black}{\noindent\textbf{Comparison:} The MOST dataset is designed to assist the recognition of multiple object states for a \textit{single object category per video}. In contrast, our proposed Ego4D-OSCA dataset focuses on anticipating the \textit{state change class of multiple objects within each video}. This means that while MOST aims to recognize objects' current state, Ego4D-OSCA emphasizes predicting what an object's state change will be as a result/effect of the next (near future), yet unobserved, action. Additionally, Ego4D-OSCA covers a larger dataset with 1,498 videos and 478 object classes, compared to MOST's 61 videos and 6 object categories. We see potential in bridging this gap in future work by adding annotations for state change classes to the MOST dataset, which could open up new avenues for research in multi-label object state change anticipation.
Additionally, contrary to the proposed Ego4D-OSCA dataset, the MOST dataset does not provide annotations regarding the actions and activities across the video.
% in which each focused object participates and the state changes it imposed due to the execution of specific actions in each activity. 
Such information, which our proposed methodology utilizes, can be crucial for understanding the progression of actions and their effects on objects enabling models to predict future states and state changes more accurately. Without this contextual information, it becomes significantly more challenging to infer the relationships between actions, object interactions, and subsequent state changes, limiting the ability to anticipate how an object might evolve within the dynamic environment of a procedural activity.}

\noindent \textbf{HowToChange~\cite{xue2023learning}:} This dataset is generated using a subset of The Food \& Entertaining category of the HowTo100M dataset~\cite{miech19howto100m}, a large-scale dataset of narrated videos. The reasons for selecting that particular subset are: (1) it constitutes one-third of the entire HowTo100M video collection, (2) cooking tasks within this category provide a rich variety of objects, tools, and state changes, making it an excellent testing ground for open-world Object-State-Change (OSC) understanding, and (3) in cooking activities, a single state transition can often be linked to a diverse array of objects, creating opportunities for compositional learning. 
The dataset provides annotation data related to three OS classes (initial, transitioning, and end state) related to OSC localization.
A total of $498,475$ videos and $11,390,287$ ASR transcriptions processed with LLAMA2 reveal the most frequently observed state transitions and the associated objects. 
This information is utilized to establish an OSC vocabulary, identifying 134 objects, 20 state transitions, and 409 unique OSCs.

\noindent\textbf{Comparison:} The HowToChange dataset contains clips that involve a single state-changing action that is not compatible with the requirement for subsequent actions in a single video as in Ego4D-OSCA. On the other hand, it contains novel objects in the test set which sets the dataset a challenging benchmark for OSC analysis. 
\vspace*{0.2cm}

% \noindent \textbf{OSDD~\cite{gouidis2021detecting}:} The Objects States Detection Dataset comprises images portraying common household objects in various states. The ground-truth annotations encompass labels and bounding boxes for 18 object categories and 9 state classes. The object categories include bottle, jar, tub, book, drawer, door, cup, mug, glass, bowl, basket, box, phone, charger, socket, towel, shirt, and newspaper. The 9 state classes cover open, close, empty, containing something liquid (CL), containing something solid (CS), plugged, unplugged, folded, and unfolded.
% \textbf{Comparison:}This dataset comprises images, unlike Ego4D-OSCA, which consists of videos. It aims to localize Object State (OS) in images, whereas Ego4D-OSCA focuses on predicting the Object State Change (OSC) in the forthcoming, unseen video segments.

\noindent \textbf{VSCOS~\cite{yu2023video}:} This dataset comprises $1,905$ video clips of an average duration of $7.4$ seconds, capturing various interactions with objects and state changes. These videos encompass 30 action categories and 124 object categories, resulting in $271$ valid combinations in total. The state changes in the dataset can be categorized into four prominent groups: Rigid Object Composition and Decomposition (e.g., combine, cut, split, disintegrate, unpackage), Non-rigid Object Transformation (e.g., pour (liquid), crack (egg)), Object Appearance Change (e.g., cook, clean), and Object Articulation (e.g., open, close, twist).

\noindent\textbf{Comparison:} Each video in the VSCOS dataset contains a single state-changing action. The test set of this dataset is challenging because it encompasses the following cases: novel objects - seen state changes, seen objects - novel state changes, and novel objects - novel state changes.
\vspace*{0.2cm}

%%%% This dataset contains 51 state-verbs. are thesee considered Obj state classes?
\noindent \textbf{VOST~\cite{Tokmakov_2023_CVPR}:} VOST is introduced as a benchmark for video object segmentation, emphasizing intricate object transformations. In contrast to current datasets, VOST introduces scenarios where objects undergo processes such as breaking, tearing, and molding, leading to substantial alterations in their overall appearance. Comprising over 700 high-resolution videos captured in diverse environments, each video in the dataset has an average duration of $21$ seconds and is meticulously labelled with instance masks of objects across frames.

\noindent\textbf{Comparison:} 
The VOST dataset comprises videos depicting objects undergoing state-changing transformations. While a change in an object's state is demonstrated in each clip of the dataset, the state types are not explicitly labelled. According to the authors, the transformations are indicated by 51 specific verbs such as cut, peel, apply, break, open, scoop, fold, mold, etc. This dataset excludes videos without any transformation. In contrast, Ego4D-OSCA consists of sequential videos that exhibit a series of object state changes alongside videos lacking such changes. Ego4D-OSCA encompasses 117 verbs and 475 objects.

\begin{figure}[t]
    \centering
    \includegraphics[width=\linewidth]{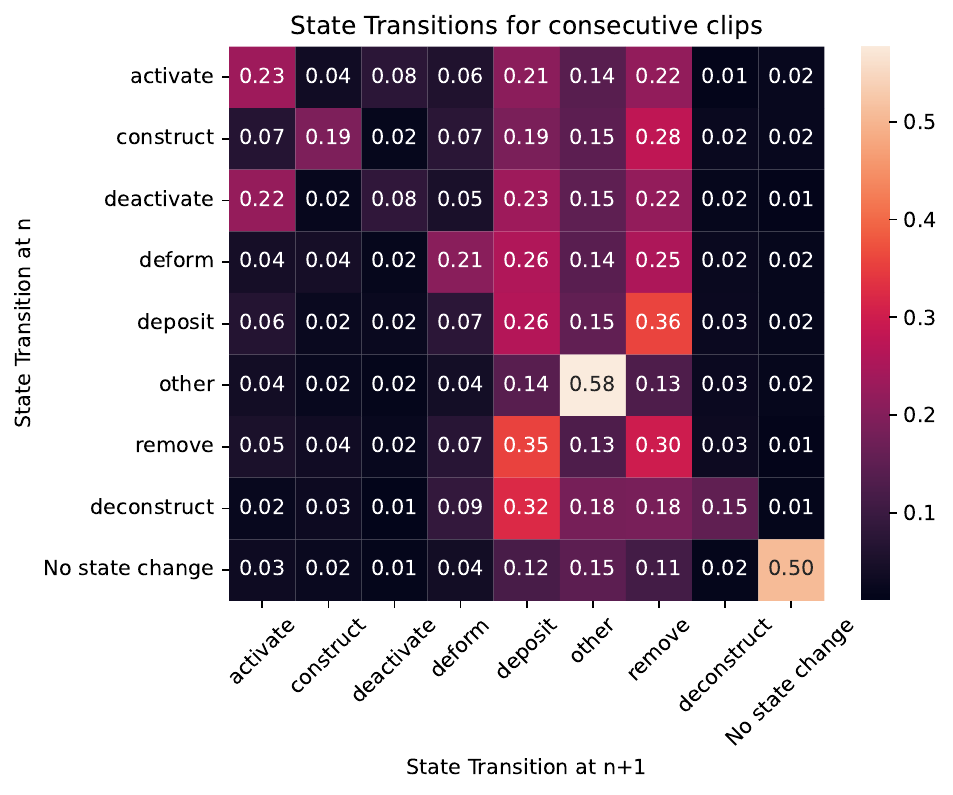}
\vspace*{0.1cm}
    \caption{Transition frequencies for all pairs of subsequent object state changes in the Ego4D-OSCA dataset videos.}
    
    \label{fig:state_pairs}
\end{figure}

\section{Ego4D-OSCA dataset statistics}
\label{sec:analysis}
% Filippos do your magic here

% \subsection{Action-Dependent Object States and State-Dependent Object Actions}

We focus on the dynamic nature of object interactions in the dataset by extracting statistics for the combinations of object states in conjunction with action verbs and object classes. 
We investigate how different action verbs are associated with a variety of object states, highlighting the diversity and context-dependence of an action's effects. 
Additionally, we examine the variability of object states based on the actions performed, emphasizing the challenges imposed for the tasks of action and activity recognition and anticipation, and object state classification and anticipation. This analysis underscores the richness of the dataset and the sophisticated modelling required to accurately interpret and predict object states in various human activity contexts.

\subsection{Action verbs vs object states}
Each histogram illustrated in Fig.~\ref{fig:states_actions_histograms1} demonstrates the distribution of the occurrences of action verbs in action segments of the proposed dataset associated with an object state change class, with challenging long-tail distributions.
The high variability of action verbs and the state change class associations are observed for the `activate', `deactivate', `construct', `deconstruct', `deposit', and `remove'.

Respectively, in Fig.~\ref{fig:states_objects_histograms2}, each histogram shows the (frequency) occurrences distribution of the instances of object classes in action segments associated with an object state change class of the proposed dataset.

In Fig.~\ref{fig:St_Verb}, the histogram provides the number of distinct object state change classes associated with various action verb classes. It can be verified that the majority of action classes are associated with at least three distinct object state classes. For example, actions involving the verb ``open" (leftmost label on the x-axis) can lead to any of the state change classes depending on the object, e.g. the ``activate" state change occurs when opening a microwave and ``deposit" when opening a box. 
The observed diversity highlights the complexity and context-dependence of state-modifying actions in the dataset capturing a wide range of interactions and their state changes on interacting objects, which makes it valuable for training and testing sophisticated predictive models. 
% In other words, this observation indicates that action verbs encompass a broad spectrum of possible state changes, which is crucial for applications in robotics, AI, and natural language processing where predicting accurately the outcomes of ongoing or near-future actions is important. 
This, in turn, indicates the need for elaborate learning models that can cope with the wide range of specific visual and semantic contexts in combination with different object classes involved in each action. 
% Furthermore, this characteristic suggests that the dataset is rich and comprehensive, 

\subsection{Objects vs object states}

The histogram in Fig.~\ref{fig:St_Obj}, illustrates that certain objects in \textit{Ego4D-OSCA} can appear in up to eight different states, depending on the action performed, which reveals the action-dependent variability of object states within the dataset. 
% This characteristic highlights the dynamic and challenging nature of object interactions and the task of object state changes classification and anticipation. This variability also poses significant challenges for action recognition, as models must accurately differentiate between state transitions caused by different actions. 
% For activity recognition, this complexity is compounded by the need to track and interpret state changes over time, requiring robust temporal models and extensive training data. 
% Additionally, object-agnostic state recognition methods face difficulties due to the necessity of recognizing states without specific object context, demanding advanced feature extraction techniques. Consequently, this dataset characteristic underscores the need for sophisticated algorithms and high-quality, context-rich training data to address the inherent complexity in recognizing and predicting action, activity, and object states accurately making a challenging dataset for methods tackling these problems.
This observation underscores the complexity and dynamic nature of object interactions as well as the challenges to be tackled by solutions for classifying and predicting object state changes.
The variability in these interactions presents significant challenges also for action recognition, where models need to accurately identify state changes and their transitions induced by subsequent actions, necessitating robust temporal models and comprehensive training data.

\begin{figure*}[t]
    \centering
    \includegraphics[width=0.95\textwidth]{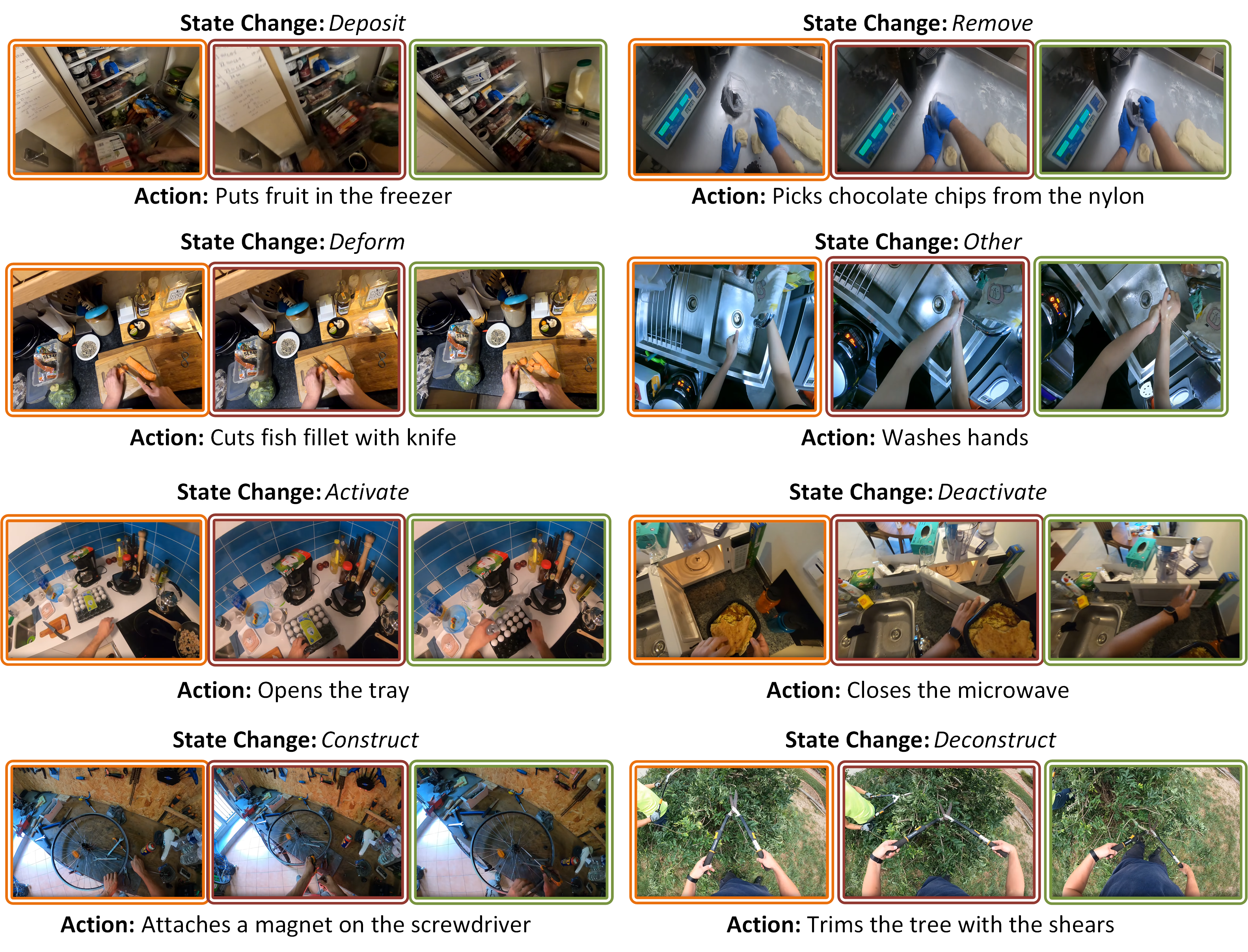}
    \vspace*{0.1cm}
    \caption{Sample frames of the Ego4D-OSCA dataset depict the initial state, the PNR frame, and the final state for the eight object state change classes (the class `No object state change' is not included). The variability of visual environments/contexts, actions, and objects associated with the object state changes classes is highlighted.}
    \label{fig:state_images}
\end{figure*}

\subsection{Variability in state changes \& activity duration}
The histogram in Fig.~\ref{fig:St_Vid_100} illustrates the distribution of state transitions observed within the first 100 videos from the dataset.
%, which contains a total of 1055 training videos (555 in the validation set)
Each bar represents the frequency of specific state transitions between actions within a video, providing insights into the temporal dynamics and complexity of activities performed in the dataset.

The histogram highlights a significant variation in the number of state transitions observed within each video sample, indicating varying levels of complexity and duration in the actions performed. For instance, video 14 (video sample: \textit{1e5bd816-e1dd-43d3-8709-42c83114dc7c}) stands out with 880 object state transitions and a duration of approximately 3 hours. This underscores the intricate nature of the actions captured in the original Ego4D dataset~\cite{grauman2022ego4d}, where the extent of state transitions, as presented in the proposed variant (\textit{Ego4D-OSCA}), reflects not only the complexity of the activities but also their temporal duration. Such variability emphasizes the need for comprehensive modelling approaches capable of accommodating diverse activity durations and complexities within the dataset. Moreover, this variability necessitates the consideration of a large action and state history by methods that utilize this information to predict future actions or object states, ensuring robustness and accuracy in forecasting.

\begin{table*}[h]

    \centering
\resizebox{\textwidth}{!}{
 \begin{tabular}{l|cc|cc|cc|c|c} \hline
      \bf OSC & \bf \rotatebox{0}{activate} & \bf \rotatebox{0}{ deactivate} & \bf \rotatebox{0}{deposit}& \bf \rotatebox{0}{remove}  & \bf \rotatebox{0}{construct} & \bf \rotatebox{0}{deconstruct}& \bf \rotatebox{0}{deform} &   \bf \rotatebox{0}{other}    \\ \hline
      \bf Pre    &    \bf \rotatebox{0}{\textcolor{violet}{pre} }   &     \bf \rotatebox{0}{\textcolor{teal}{pre}} &  \bf \rotatebox{0}{\textcolor{orange}{pre}} &  \bf \rotatebox{0}{\textcolor{olive}{pre}} &   \bf \rotatebox{0}{\textcolor{pink}{pre}} &  \bf \rotatebox{0}{\textcolor{purple}{pre}} &    \bf \rotatebox{0}{pre} &    \bf \rotatebox{0}{pre}    \\  
      \bf      &    \bf \rotatebox{0}{\textcolor{violet}{activate} }   &     \bf \rotatebox{0}{\textcolor{teal}{deactivate}} &  \bf \rotatebox{0}{\textcolor{orange}{deposit}} &  \bf \rotatebox{0}{\textcolor{olive}{remove}} &   \bf \rotatebox{0}{\textcolor{pink}{construct}} &  \bf \rotatebox{0}{\textcolor{purple}{deconstruct}} &    \bf \rotatebox{0}{deform} &    \bf \rotatebox{0}{other}    \\  \hline      
    \bf Post  &   \bf \rotatebox{0}{\textcolor{teal}{post}}  &
 \bf \rotatebox{0}{\textcolor{violet}{post}}  &
                 \bf \rotatebox{0}{\textcolor{olive}{post}} 
  &     \bf \rotatebox{0}{\textcolor{orange}{post}} 
      &      \bf \rotatebox{0}{\textcolor{purple}{post}}
 &     \bf \rotatebox{0}{\textcolor{pink}{post}} 
  &     \bf \rotatebox{0}{post} &
    \bf \rotatebox{0}{post} \\ 

    \bf  &   \bf \rotatebox{0}{\textcolor{teal}{activate}}  &
 \bf \rotatebox{0}{\textcolor{violet}{deactivate}}  &
                 \bf \rotatebox{0}{\textcolor{olive}{deposit}} 
  &     \bf \rotatebox{0}{\textcolor{orange}{remove}} 
      &      \bf \rotatebox{0}{\textcolor{purple}{construct}}
 &     \bf \rotatebox{0}{\textcolor{pink}{deconstruct}} 
  &     \bf \rotatebox{0}{deform} &
    \bf \rotatebox{0}{other} \\ \hline

    \end{tabular}
    }
    \caption{The super-annotated state change labels and the corresponding pre-/post-state labels of a video segment, where the state modifying action occurs. The pairs activate-deactivate, deposit-remove, and construct-deconstruct constitute pairs of inverse state change actions. Frame state labels that correspond to the same state are depicted with the same colour.}
    \label{tab:states_annots}

\end{table*}

Building on the significant diversity in the number of state transitions observed within each video, the transition matrix in Fig.~\ref{fig:state_pairs} provides further insight into the probabilistic nature of these transitions. For instance, the data indicate that when an object is in the `activate' state during action $n$, it frequently transitions to the `deposit' or `remove' states in action $n+1$. This suggests that actions involving the activation of objects, typically electrical appliances, are usually followed by actions that involve placing items into or removing items from these objects. Such logical transitions underscore the presence of temporal action ordering and causality in activities, where one action sets the stage for subsequent actions. This pattern highlights the importance of understanding state persistence and transitions, in developing predictive models. Accurately capturing these probabilistic dependencies is essential for models to effectively anticipate future states and actions. These insights reinforce the necessity for models to consider extensive action and state histories to accommodate the intricate and dynamic nature of the actions and activities in the Ego4D dataset, and inherently also in the proposed \textit{Ego4D-OSCA}.

%- Example images of ambiguous cases? where a frame-wise Object State Recognizer might fail and thus temporal associations should be considered.

\section{On the super-annotation of object state change classes}
\label{Sec: ObjectState_general}
\begin{comment}
The original Ego4D dataset does not include annotations for the specific state labels of individual video frames. Instead, annotations about state changes are provided, which relate to entire video segments. Additionally, the dataset includes annotations for bounding boxes and object classes across seven critical frames within each video segment. These frames are temporally centered around the occurrence of the state change that occurs within each video segment. Based on this information, we super-annotate certain critical frames of each video segment with state-related labels as follows. For each video segment, we annotate the initial and final frames as $pre\_X$ and $post\_X$, respectively, where $X$ denotes the label of the state change. Furthermore, in line with the semantic implications of these changes, we establish three pairs of state changes. Each pair is constructed under the premise that the first action is the inverse of the second concerning the resulting state change. For instance, if $X$ and $Y$ represent inverse state changes, then the labels $pre\_X$ and $post\_Y$ are considered samples of identical states. A similar correspondence applies between $pre\_Y$ and $post\_X$. For example, the states $pre\_remove$ and $post\_deposit$ are considered identical, since $remove$ and $deposit$ constitute a pair of inverse state changes. Table~\ref{tab:states_annots} delineates the specifics of these super-annotated states. 
\end{comment}

As stated in the main paper, the original Ego4D dataset does not provide specific state labels for individual video frames; instead, it offers annotations on state changes tied to entire video segments. These annotations include object bounding boxes and classes for seven key frames, centered around the moment of state change in each segment. Building on this, we augment critical frames with state-related labels. Specifically, for each segment, we label the initial frame as $pre\_X$ and the final frame as $post\_X$, where $X$ represents the state change. To capture the semantics of these state transitions, we define three pairs of inverse state changes. Each pair reflects that one action reverses the outcome of the other. For instance, if $X$ and $Y$ are inverse changes, then $pre\_X$ and $post\_Y$ are considered equivalent, as are $pre\_Y$ and $post\_X$. A practical example of this is the pair $pre\_remove$ and $post\_deposit$, since "remove" and "deposit" are inverse actions. Table~\ref{tab:states_annots} outlines these super-annotated state labels and their inverse association.

\begin{figure}[h]
    \centering
    \includegraphics[width=\columnwidth]{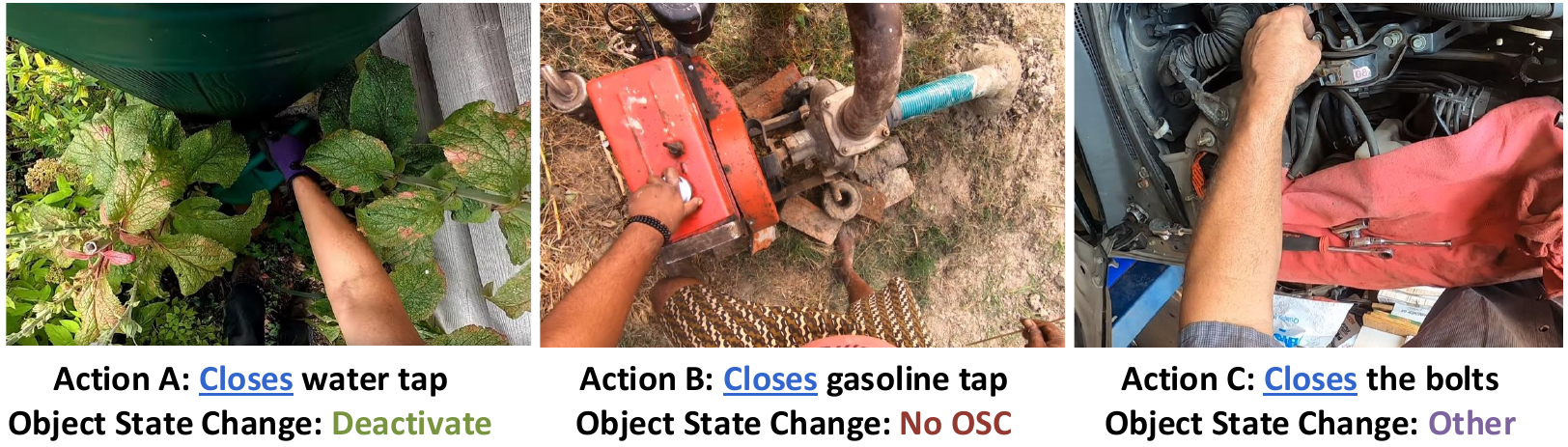}
    \caption{Sample frames from 3 instances of the ``close” action, each involving different contexts and objects from the Ego4D-OSCA dataset, each resulting in various types of state changes.} 
    \label{fig:fig5}
\end{figure}

To emphasize the complexity and context variability of state change classes, Figure~\ref{fig:state_images} shows sample images from the Ego-OSCA dataset depicting the object state change visual progression.
In those few samples, one may notice the inverse association between different state change stages, as well as the large variability of visual environments and contexts, actions, and objects involved in different classes of object state changes.
\textcolor{black}{Finally, as also stated in lines $84-86$ and shown in Fig. 3 of the main paper, it is worth mentioning that motion motifs (as defined by verb primitives) do not necessarily have a one-to-one correspondence with states. As an example, in Fig.~\ref{fig:fig5} we observe that the verb ``close” can result to more than one object state change class. This highlights the fact that the object-related context is as important as the motion motif when defining an action as well as when estimating the anticipated object state change due to the execution of the action. Therefore to address the OSCA task an ideal method should build upon the past and current estimates of object detection and state estimation, as well as action recognition methods in order to robustly estimate the anticipated state of an object in procedural activities. }

% \clearpage

\begin{figure*}[t]
    \centering
   % \includegraphics[width=\textwidth]{images/hist12.pdf}
 %  \begin{subfigure}{.5\textwidth}
\subfloat{{\includegraphics[width=.5\textwidth]{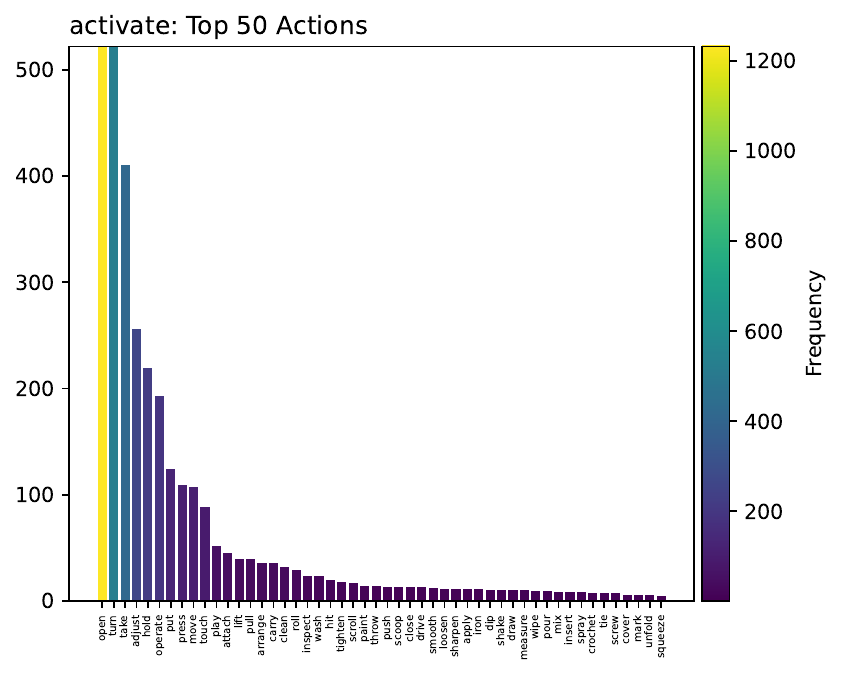} }}%  
\subfloat{{\includegraphics[width=.5\textwidth]{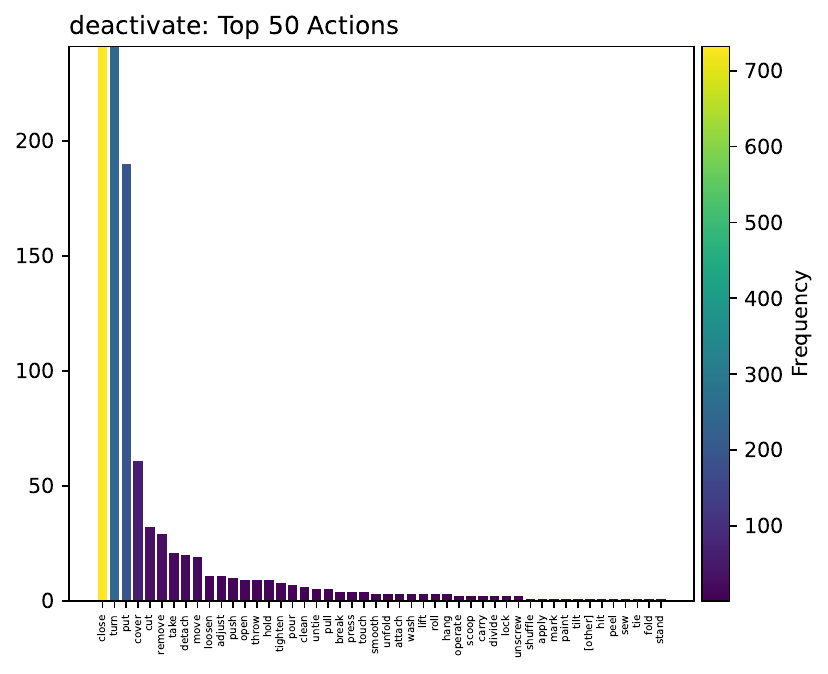} }}%
% \end{figure}
% \begin{figure}
  % \begin{subfigure}{.5\textwidth}

%\end{subfigure}
\subfloat{{\includegraphics[width=.5\textwidth]{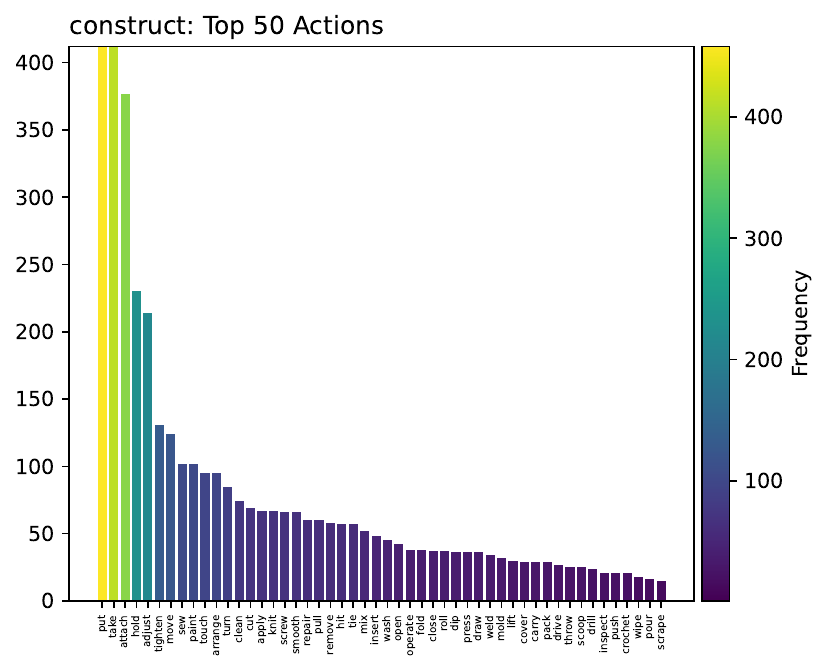}}}
\subfloat{{\includegraphics[width=.5\textwidth]{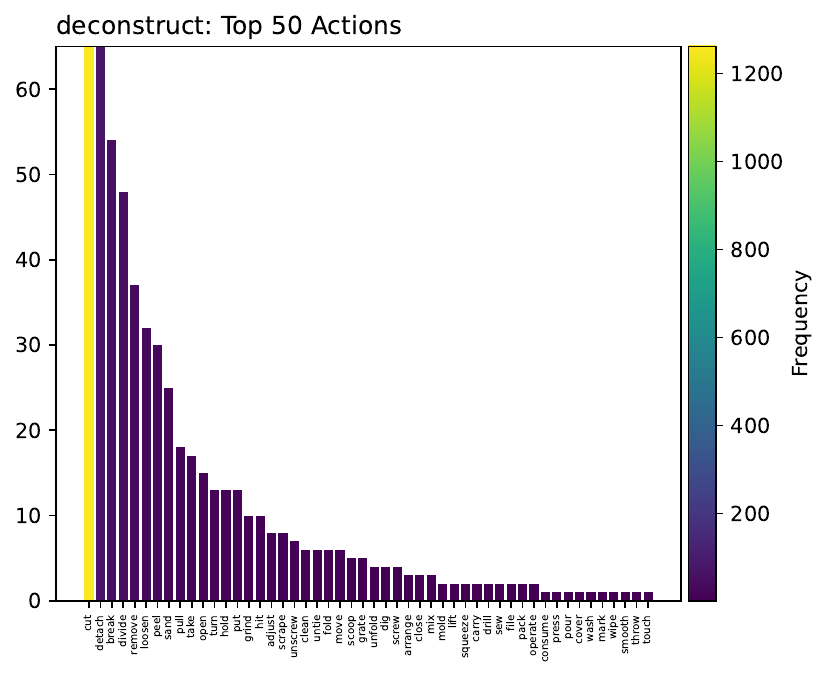}}}

\subfloat{{\includegraphics[width=.5\textwidth]{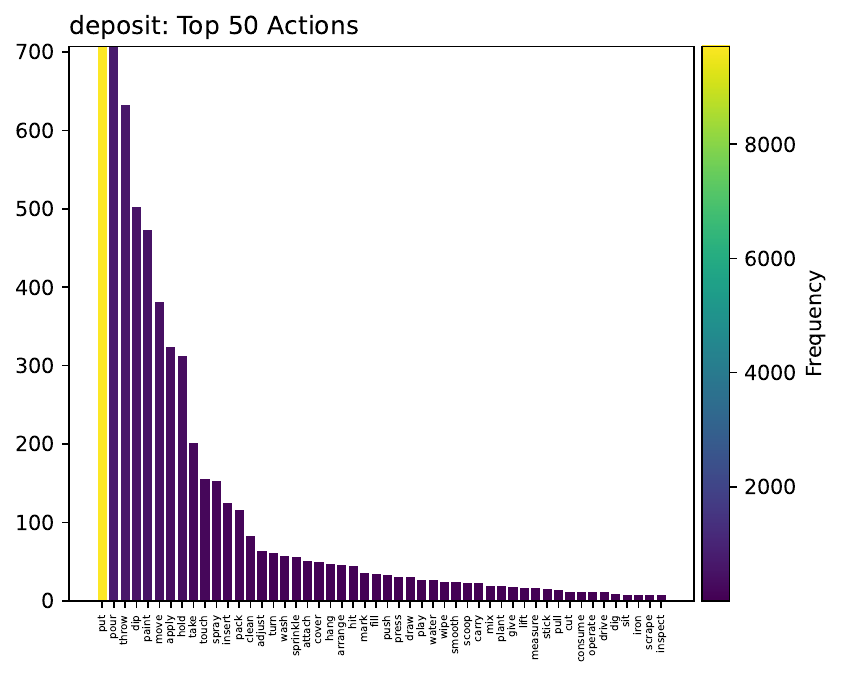}}}
\subfloat{{\includegraphics[width=.5\textwidth]{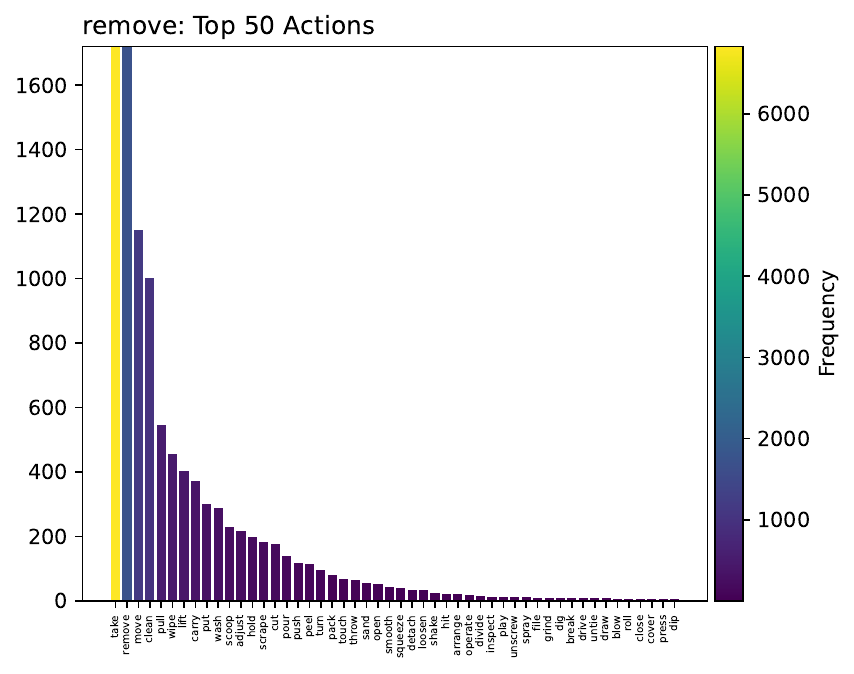}}}
 \vspace*{0.1cm}  
    \caption{The frequency distribution of the top 50 actions (occurrences of action classes in the dataset action segments) concerning an object state change class is illustrated in each histogram for the classes `activate', `deactivate', `construct', `deconstruct', `deposit', `remove'.}
    \label{fig:states_actions_histograms1}
\end{figure*}

   \begin{figure*}
    \centering
   % \includegraphics[width=\textwidth]{images/hist12.pdf}
 %  \begin{subfigure}{.5\textwidth}
\subfloat{{\includegraphics[width=.5\textwidth]{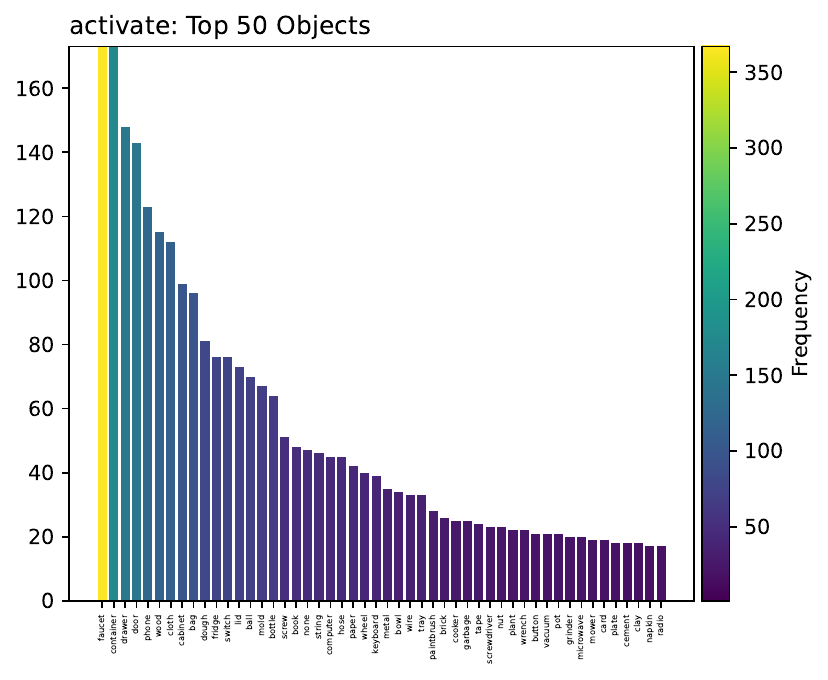} }}%  
\subfloat{{\includegraphics[width=.5\textwidth]{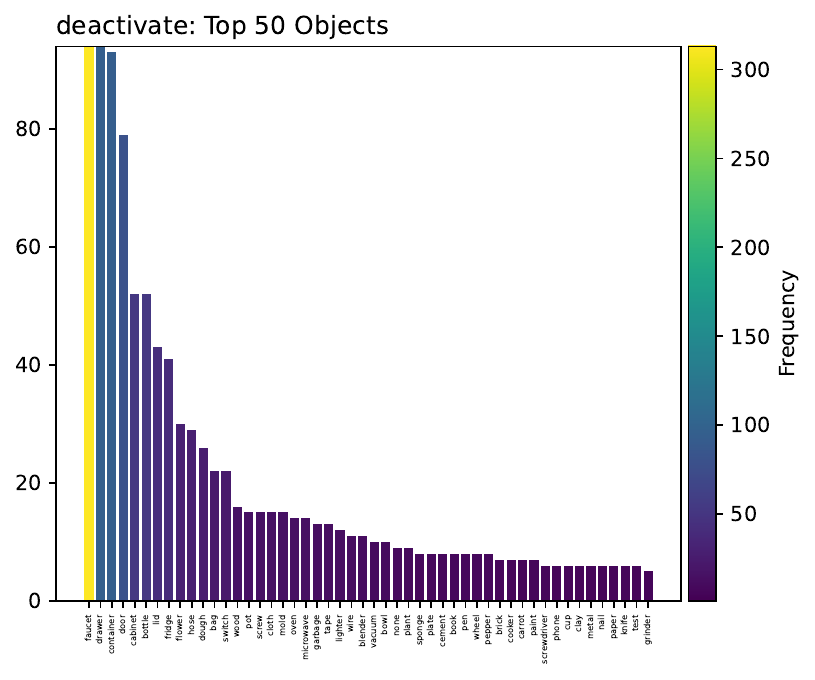} }}%
% \end{figure}
% \begin{figure}
  % \begin{subfigure}{.5\textwidth}

%\end{subfigure}
\subfloat{{\includegraphics[width=.5\textwidth]{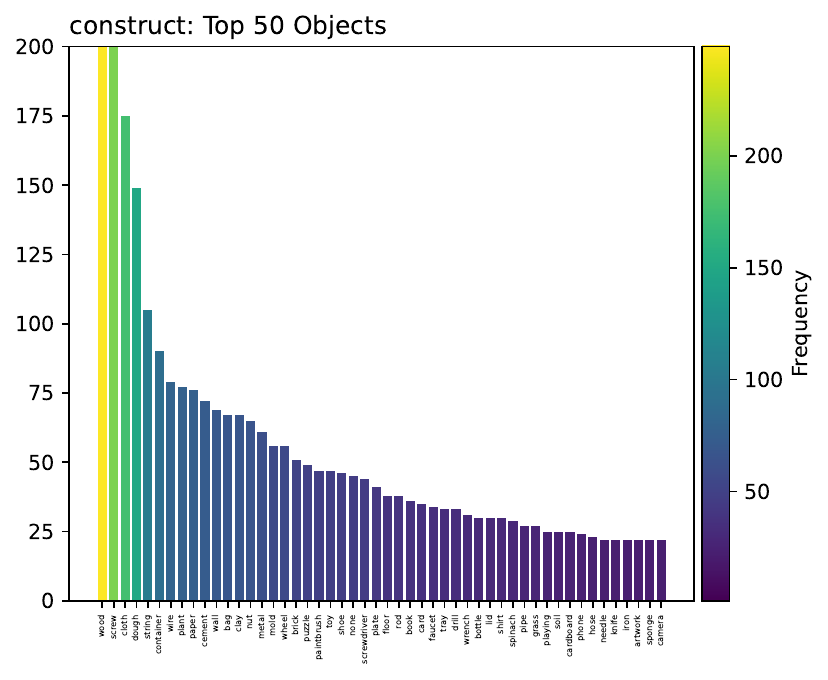}}}
\subfloat{{\includegraphics[width=.5\textwidth]{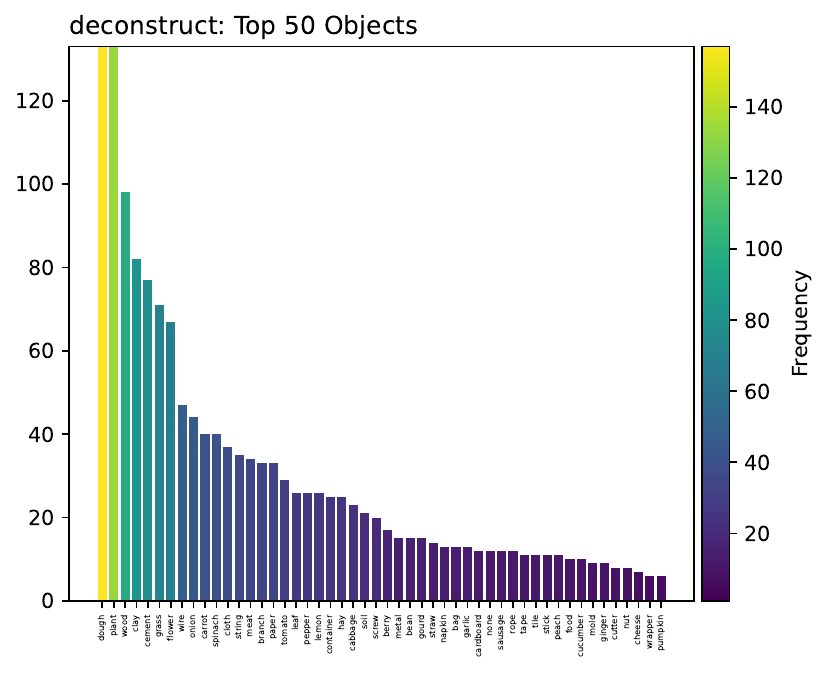}}}

\subfloat{{\includegraphics[width=.5\textwidth]{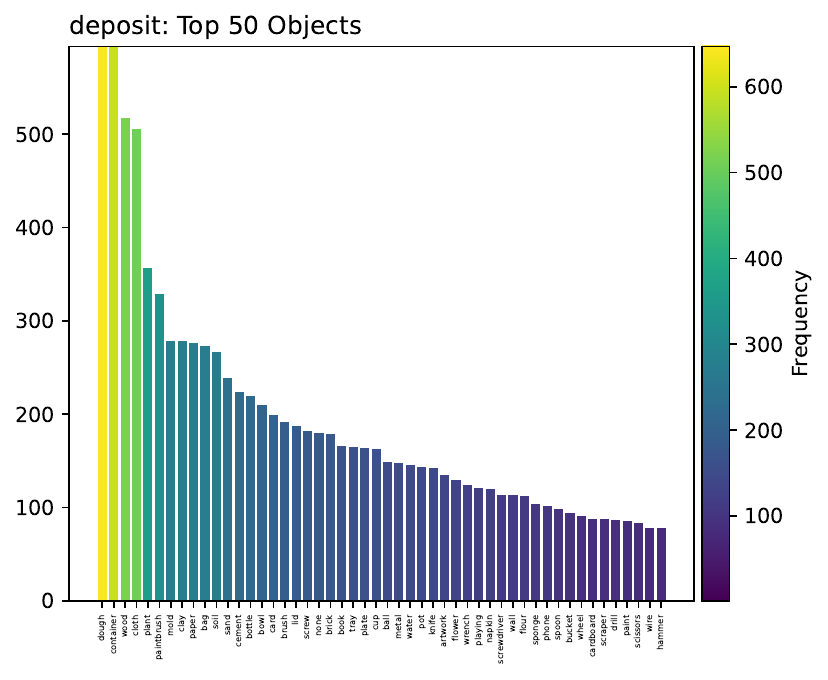}}}
\subfloat{{\includegraphics[width=.5\textwidth]{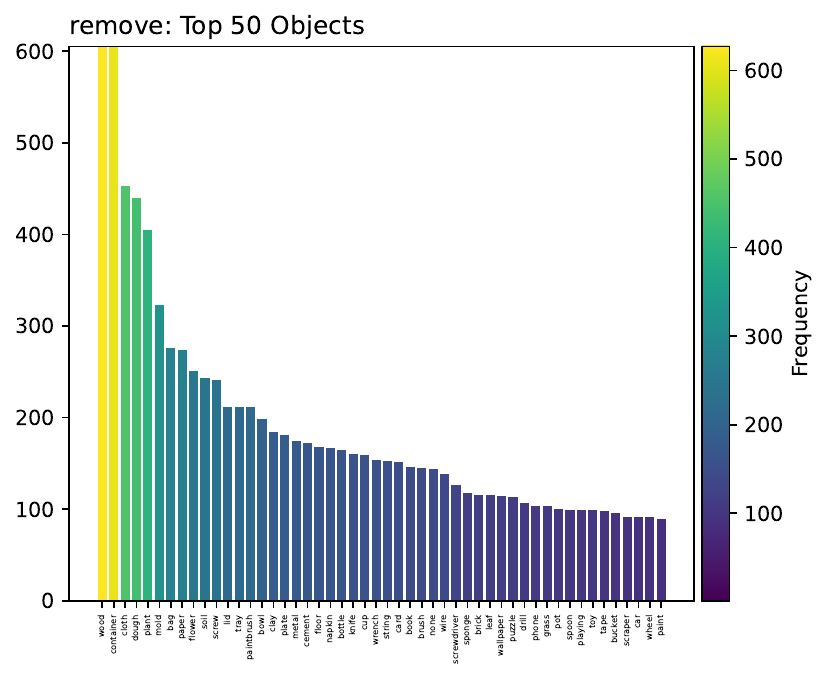}}}
\vspace*{0.1cm}
    \caption{The frequency distribution of the top 50 objects (occurrences of object classes based on the dataset action segments) concerning an object state change class is illustrated in each histogram for the classes `activate', `deactivate', `construct', `deconstruct', `deposit', `remove'.}
    \label{fig:states_objects_histograms2}
\end{figure*}

\begin{figure*}[t]
    \centering
    \includegraphics[width=\textwidth]{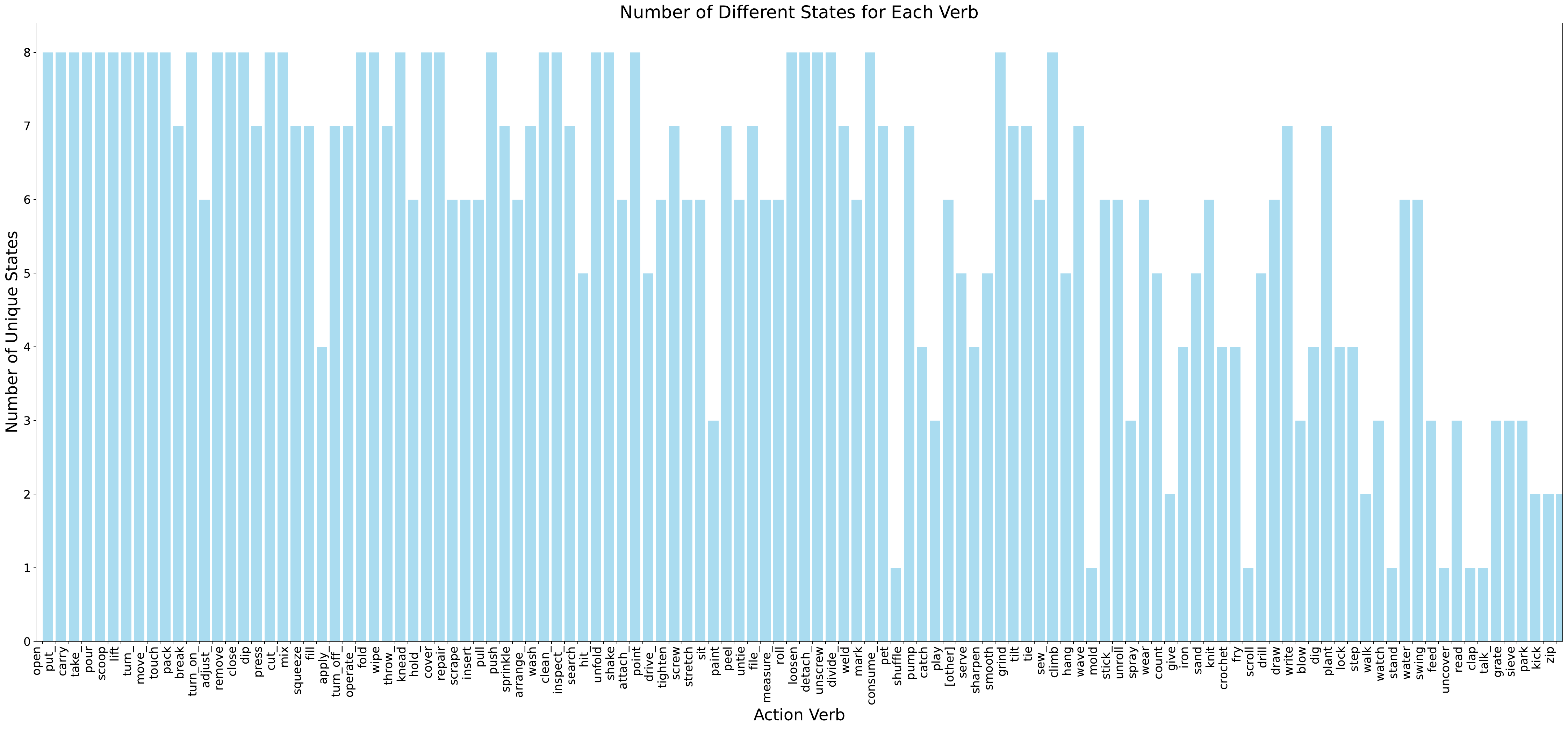}
    \vspace*{0.1cm}
    \caption{Histogram of object states associated with action verb classes.}
    \label{fig:St_Verb}
\end{figure*}

\begin{figure*}[t]
    \centering
    \includegraphics[width=\textwidth]{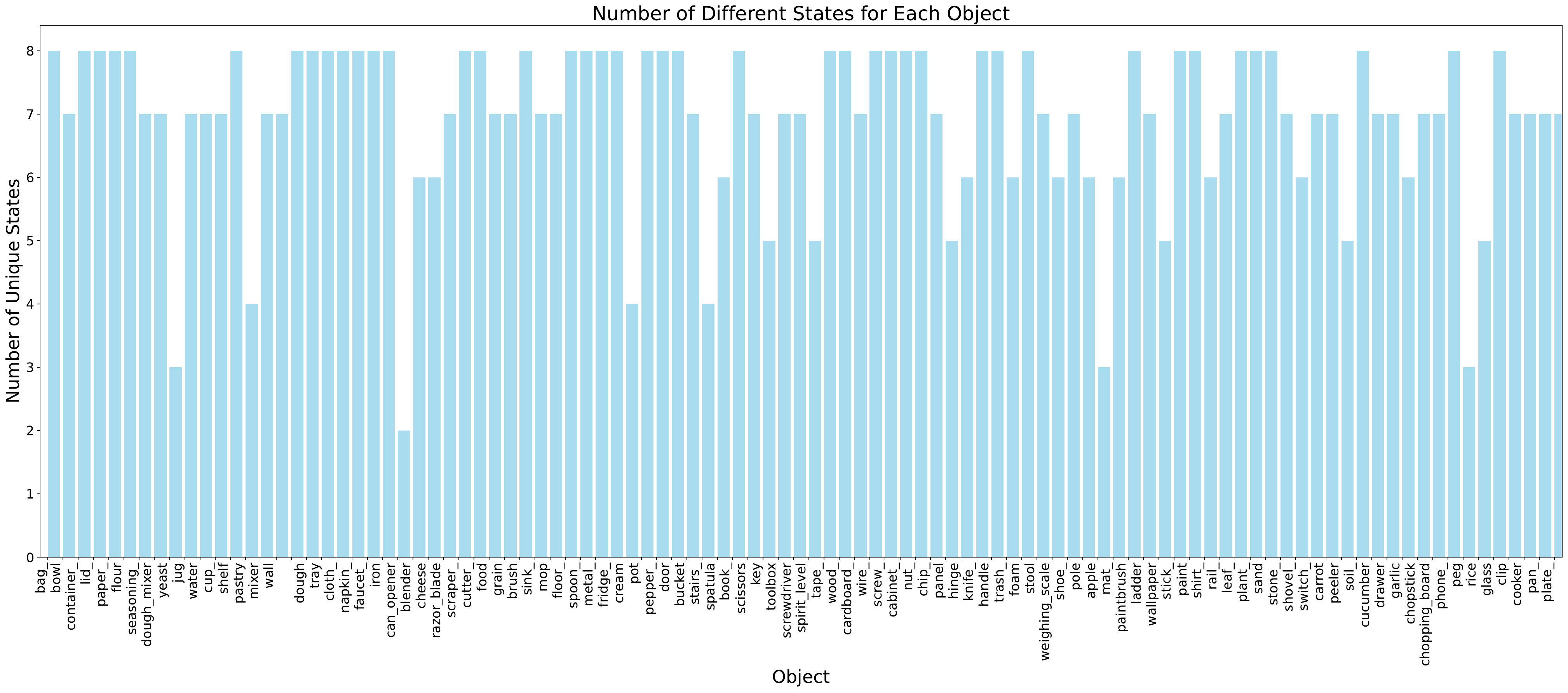}
    \vspace*{0.1cm}
    \caption{Histogram of object states associated with object classes. For better visualization purposes, we only depict the variability in the states of the first 100 objects.}
    \label{fig:St_Obj}
\end{figure*}

\begin{figure*}[t]
    \centering
    \includegraphics[width=\textwidth]{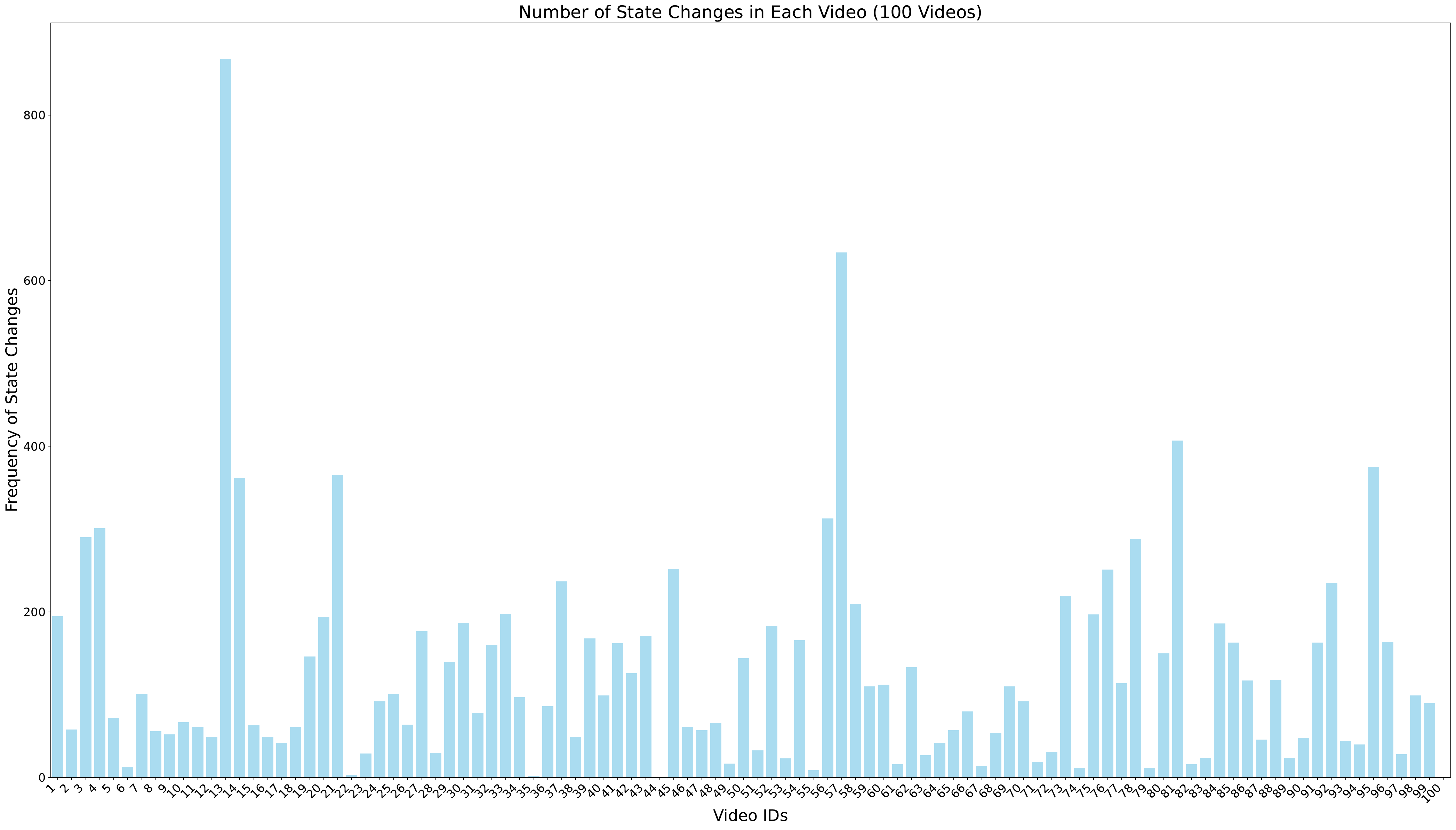}
    \vspace*{0.1cm}
    \caption{Histogram of the frequency of state transitions in the first 100 videos.}
    \label{fig:St_Vid_100}
\end{figure*}

% \section{Rationale}
% \label{sec:rationale}
% % 
% Having the supplementary compiled together with the main paper means that:
% % 
% \begin{itemize}
% \item The supplementary can back-reference sections of the main paper, for example, we can refer to \cref{sec:intro};
% \item The main paper can forward reference sub-sections within the supplementary explicitly (e.g. referring to a particular experiment); 
% \item When submitted to arXiv, the supplementary will already included at the end of the paper.
% \end{itemize}
% % 
% To split the supplementary pages from the main paper, you can use \href{https://support.apple.com/en-ca/guide/preview/prvw11793/mac#:~:text=Delete%20a%20page%20from%20a,or%20choose%20Edit%20%3E%20Delete).}{Preview (on macOS)}, \href{https://www.adobe.com/acrobat/how-to/delete-pages-from-pdf.html#:~:text=Choose%20%E2%80%9CTools%E2%80%9D%20%3E%20%E2%80%9COrganize,or%20pages%20from%20the%20file.}{Adobe Acrobat} (on all OSs), as well as \href{https://superuser.com/questions/517986/is-it-possible-to-delete-some-pages-of-a-pdf-document}{command line tools}.


\end{document}